%% file: main.tex
\documentclass{article}
\PassOptionsToPackage{table}{xcolor}
\usepackage{iclr2027_conference,times}
\iclrfinalcopy

\input{math_commands.tex}

\usepackage{hyperref}
\usepackage{marvosym}

\usepackage{xspace}
\usepackage{multirow}
\usepackage{xcolor}
\usepackage{tabularx}
\usepackage{pifont}
\usepackage{subcaption}
\usepackage{amssymb}
\usepackage{graphicx}
\usepackage{booktabs}

\newcommand{\sota}{\textbf}
\definecolor{navyblue}{HTML}{0071BC}

\newcommand{\cmark}{\ding{51}}
\newcommand{\cmarkg}{\textcolor{lightgray}{\ding{51}}}
\newcommand{\xmark}{\ding{55}}
\newcommand{\xmarkg}{\textcolor{lightgray}{\ding{55}}}

\makeatletter
\DeclareRobustCommand\onedot{\futurelet\@let@token\@onedot}
\def\@onedot{\ifx\@let@token.\else.\null\fi\xspace}

\makeatother

\newcommand{\name}{GeoThinker-\emph{F}\xspace}

\title{\name: Endogenous Predictive World Representations for Fast Spatial Intelligence}

\author{
Haoyuan Li$^{*1\dagger}$, Qihang Cao$^{*2}$, Huanyuan Zhou$^{3}$, Xin Hu$^{3}$, Sijia Chen$^{3}$, Zihan Guo$^{1}$, Kun Xiang$^{1}$,\\[5pt] \textbf{Yunshuang Nie}$^{1}$, \textbf{Tao Tang}$^{1}$, \textbf{Xinlong Zheng}$^{3}$, \textbf{Jia-Wang Bian}$^{2}$, \textbf{Xianming Liu}$^{3}$, \textbf{Xiaodan Liang}$^{1,4}$\textsuperscript{\Letter}\\[5pt]
$^1$Shenzhen campus of Sun Yat-sen University,
$^2$Nanyang Technological University, \\[5pt]
$^3$XPeng Motors,
$^4$MBZUAI, \\[5pt]
\url{https://github.com/Li-Hao-yuan/GeoThinker}.
}

\begin{document}

\renewcommand{\thefootnote}{$\ast$}
\footnotetext{Equal contribution and \textsuperscript{$\dagger$} Work done as an intern at XPeng Motors.}

\renewcommand{\thefootnote}{\Letter}
\footnotetext{Corresponding author.}

\maketitle
\lhead{Preprint}

\begin{center}
    \centering
    \captionsetup{type=figure}
    \begin{center}
        \includegraphics[width=\textwidth]{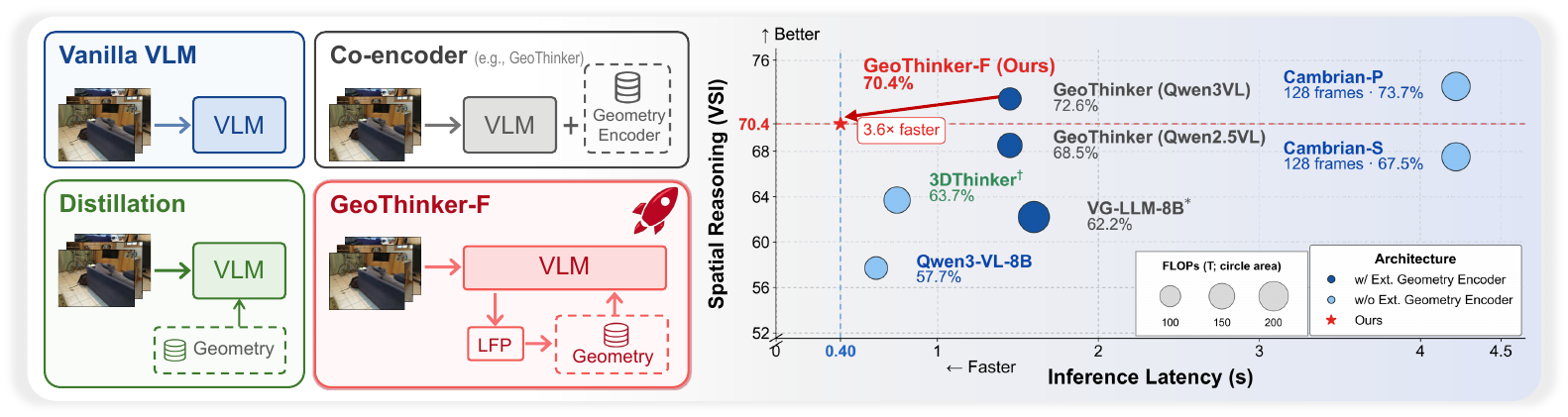}

    \end{center}
 \caption{
 \textbf{Comparison of spatial perception paradigms and their accuracy--efficiency trade-off. Left:} Existing approaches either reason from semantic visual features alone, retain an external geometry encoder at inference, or distill geometric knowledge into compact latent representations. In contrast, \name performs endogenous prediction in the MLLM latent space, transforming its causal semantic states into complementary world representations of current geometry and future semantics without executing an external geometry encoder at runtime. \textbf{Right:} With the same Qwen3-VL backbone, \name reduces inference latency from $1.45$\,s to $0.40$\,s ($3.6\times$) relative to GeoThinker while maintaining competitive spatial reasoning performance, yielding a favorable accuracy--efficiency frontier. $^{\dagger}$ denotes an optimistic backbone only lower-bound estimation.
 }
    \label{fig:motivation}
\end{center}

\input{sections/0_abstract}
\input{sections/1_intro}

\input{sections/2_related_work}
\input{sections/3_method}

\input{sections/4_experiment}
\input{sections/5_conclusion}

\bibliographystyle{iclr2027_conference}
\bibliography{iclr2027_conference}

\appendix
\input{sections/6_appendix}

\end{document}

%% file: math_commands.tex
\usepackage{amsmath,amsfonts,bm}

\def\eqref#1{equation~\ref{#1}}

\def\1{\bm{1}}

\DeclareMathAlphabet{\mathsfit}{\encodingdefault}{\sfdefault}{m}{sl}
\SetMathAlphabet{\mathsfit}{bold}{\encodingdefault}{\sfdefault}{bx}{n}



%% file: sections/0_abstract.tex
\begin{abstract}

Existing geometry-aware Multimodal Large Language Models (MLLMs) increasingly rely on external 3D encoders to provide structural priors, but executing these additional encoder paths at inference introduces substantial latency and computational overhead. We propose \name, an efficient framework that internalizes spatiotemporal perception by transforming causal MLLM hidden states into predictive latent world representations. Our key insight is that intermediate semantic representations contain sufficient information to predict useful representations of current geometry and future semantics, without requiring an exact reproduction of features from a dedicated 3D encoder. To realize this idea, \name adopts a decoupled two-stage training strategy. In the first stage, the MLLM backbone is frozen while lightweight Latent Frame Prediction (LFP) heads learn from an offline 3D geometry expert and next-frame semantic targets. In the second stage, the external expert is removed, and the model is jointly optimized to generate and selectively integrate its self-predicted priors through a specialized Spatial-Grounded Temporal (SGT) Fusion module. 
Extensive evaluations show that \name eliminates runtime dependence on external 3D geometry encoders while maintaining competitive performance across diverse spatial intelligence benchmarks. It achieves up to a $3.6\times$ inference speedup, reducing latency from $1.45\,\mathrm{s}$ to $0.40\,\mathrm{s}$ while retaining strong spatial reasoning performance.


\end{abstract}

%% file: sections/1_intro.tex
\vspace{-2mm}
\section{Introduction}
\vspace{-1mm}
Spatial intelligence has emerged as a central capability for Multimodal Large Language Models (MLLMs), with broad applications in 3D scene understanding~\citep{cai2025scaling(sensenova),yang2025cambrian,cao2026vggt(caoyang), cao2023coda}, embodied AI~\citep{zhou2025roborefer,xu20253d}, and autonomous decision-making~\citep{li2025recogdrive}. To bridge the gap between 2D semantic perception and 3D structural understanding, recent geometry-aware frameworks~\citep{zheng2025learning(vgllm),fan2025vlm(vlm3r),chen2025reasoning(gsreasoner),li2026thinking(geothinker)} increasingly leverage priors from external 3D geometry foundation models~\citep{wang2025vggt,wang2025pi3,wang2025continuous(cut3r)}, either by integrating their features into the inference pipeline or distilling their representations during training. 
By jointly processing multiple frames, these geometry models provide fine-grained structural cues that complement the semantic representations of conventional visual encoders for reliable spatial relation understanding across views.

However, methods that retain a dedicated 3D geometry encoder at inference incur substantial computation, memory consumption, and synchronization latency, limiting deployment in latency-sensitive settings. 
%
%
Training-time distillation~\citep{chen2025think(3dthinker)} avoids this dependency by absorbing geometric supervision into the MLLM parameters, but does not provide a dense, frame-aligned geometry interface for explicit routing across intermediate layers at inference. Dual-encoder architectures~\citep{li2026thinking(geothinker)} retain such an explicit feature interface, but at the cost of executing an external geometry branch.
Our key insight is that intermediate semantic representations contain sufficient information to predict useful latent representations of current geometry and future semantics, without exactly reproducing features from a dedicated 3D encoder. This raises a natural question: 
%

\begin{quote}
\itshape
Can an MLLM generate geometric and temporal priors from its own hidden states, thereby eliminating runtime dependence on external geometry models?
\end{quote}

To eliminate this runtime overhead, we introduce \name, an efficient framework that internalizes predictive latent world representations within the MLLM. Rather than executing a dedicated 3D geometry encoder at inference, \name learns to transform intermediate hidden states into complementary representations of current geometry and future semantics. We realize this design through a decoupled two-stage training strategy. In Stage 1 (LFP pretraining), the MLLM backbone is frozen, while only lightweight Latent Frame Prediction~\citep{yang2025cambrian} (LFP) MLP heads attached to intermediate layers are trained using latent geometry features from an offline 3D expert and next-frame semantic features as prediction targets. In Stage 2 (self-driven tuning), the external geometry expert is removed, and the self-predicted representations are selectively integrated into intermediate MLLM layers through a Spatial-Grounded Temporal (SGT) Fusion module for spatial reasoning. At inference, \name generates both representations endogenously within its own latent space, eliminating runtime dependence on external geometry encoders.

Together, these components shift geometry-aware spatial reasoning from runtime external feature extraction toward endogenous latent prediction. On the Qwen3-VL-8B~\citep{Qwen3-VL} backbone, the complete \name system, combined with spatial token compression, reduces inference latency from $1.45\,\text{s}$ to $0.40\,\text{s}$, yielding a $3.6\times$ speedup while achieving a VSI-Bench score of $70.4$. Across six spatial intelligence benchmarks, \name further reaches an average score of $69.29$, demonstrating strong cross-benchmark performance without executing an external geometry encoder at inference. By jointly predicting representations of current geometry and future semantics, \name integrates spatial structure and short-horizon temporal evolution within the MLLM latent space, providing an efficient alternative to runtime multi-encoder architectures. 

%
Our contributions can be summarized as follows:

\begin{itemize}
    \item \textbf{Endogenous Predictive Latent Modeling:} We propose \name, an efficient framework that predicts complementary representations of current geometry and future semantics from causal intermediate MLLM states. Through decoupled two-stage training, an external 3D geometry model is used only as an offline teacher, eliminating the need to execute a dedicated geometry encoder at inference.

    \item \textbf{Specialized Spatiotemporal Fusion:} We introduce Spatial-Grounded Temporal (SGT) Fusion, which employs separate geometry and temporal branches to selectively integrate the self-predicted representations. A layer-interleaved injection strategy further separates the two feature streams across intermediate MLLM layers, reducing interference during spatial reasoning and preserving their complementary spatial cues.

    \item \textbf{Inference Efficiency and Cross-Benchmark Performance:} Under controlled training data, \name achieves performance comparable to the dual-encoder baseline GeoThinker while removing its runtime geometry branch. Across six spatial intelligence benchmarks, \name reaches an average score of $69.29$. When combined with spatial token compression, it reduces inference latency from $1.45\,\text{s}$ to $0.40\,\text{s}$ on Qwen3-VL-8B, yielding a $3.6\times$ speedup in end-to-end execution time.
\end{itemize}

\begin{figure*}[t]
	\centering
        \includegraphics[width=0.98\textwidth]{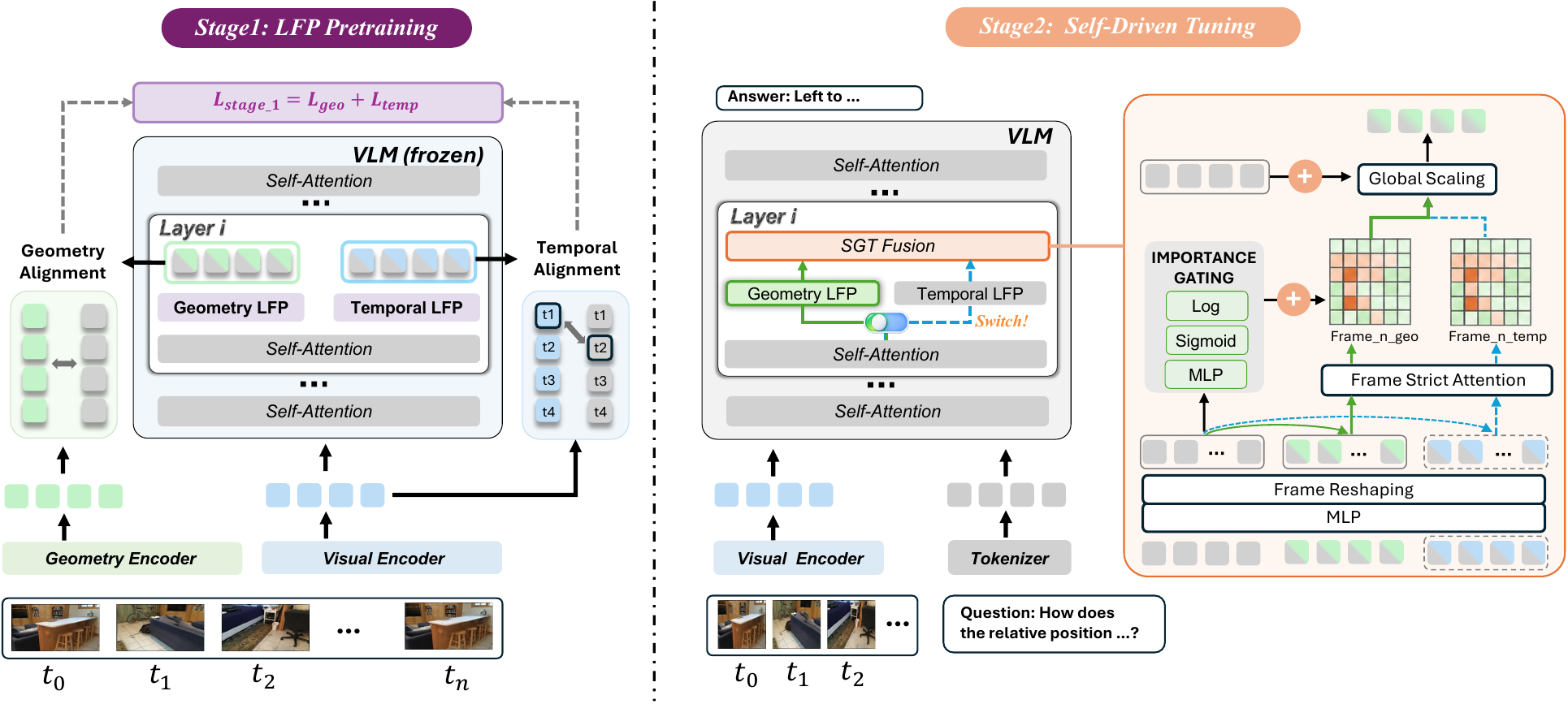}     
        \caption{
        \textbf{Overview of the \name pipeline.} The framework adopts two decoupled training stages to internalize geometric and temporal representations within the MLLM. \textbf{Stage 1 (Left):} With the MLLM backbone frozen, Latent Frame Prediction (LFP) heads attached to intermediate layers learn to predict latent geometry features produced by an offline 3D expert and semantic features of the next frame. \textbf{Stage 2 (Right):} The external geometry encoder is removed from the pipeline. 
        The resulting geometry and future-semantic representations are selectively integrated by SGT-Fusion modules inserted at intermediate MLLM layers for spatial reasoning.
        }
	\label{fig:model_overview}
    \vspace{-3mm}
\end{figure*}

%% file: sections/2_related_work.tex
\section{Related Work}
\vspace{-1mm}

\subsection{Semantics-Driven Spatial Reasoning}
MLLMs~\citep{yang2025qwen3technicalreport,team2023gemini,openai_gpt5_systemcard} have demonstrated remarkable capabilities in general visual comprehension across both static images and video sequences. Nevertheless, recent evaluations~\citep{yang2025thinking(vsi)} highlight a persistent deficiency in dependable spatial reasoning, positioning spatial intelligence as a critical hurdle in achieving human-level multi-modal reasoning. To address this limitation, existing literature has investigated several distinct methodologies.
One prominent direction involves the integration of explicit 3D information into the model pipeline. For instance, Video-3D LLM~\citep{zheng2025video(video3dllm)} enriches video tokens with frame-wise 3D spatial coordinates back-projected from RGB-D data to capture location-sensitive representations. 
Conversely, another line of work emphasizes implicit geometric regularization within the latent space. RoSS3D~\citep{wang2025ross3d} leverages cross-view and bird's-eye-view (BEV) reconstruction targets alongside denoising-style objectives to foster geometry-consistent feature spaces. 
Parallel to architectural modifications, expanding training data has emerged as a highly potent strategy. Cambrian-S~\citep{yang2025cambrian} curates the VSI-590K dataset to investigate the boundaries of data scaling in spatial contexts. Similarly, SenseNova-SI~\citep{cai2025scaling(sensenova)} introduces the massive SenseNova-SI-8M dataset, achieving substantial performance leaps on both VSI-Bench and the EASI leaderboard~\citep{cai2025holistic} without sacrificing general multimodal proficiency. 
Complementing these approaches, reasoning-centric optimization seeks to fully unlock the inherent cognitive power of the underlying LLM. SpatialLadder~\citep{li2025spatialladder} reinforces complex, multi-step spatial reasoning by utilizing reinforcement learning paired with verifiable reward signals. 
%
While these approaches avoid persistent external encoders and preserve standard inference efficiency, they typically acquire spatial structure indirectly through training data, auxiliary objectives, or reasoning supervision, rather than generating dedicated geometry representations at inference. Our work instead uses a pretrained geometry encoder as a training-time teacher, enabling the MLLM to predict geometry-derived structural priors internally without introducing an additional encoder at runtime.

\vspace{-1mm}
\subsection{Prior-Driven Geometry Integration}
\vspace{-1mm}

To endow MLLMs with spatial intelligence, recent architectural innovations have explored various pathways to incorporate 3D geometric representations from dedicated foundation models (e.g., VGGT~\citep{wang2025vggt}, $\pi^3$~\citep{wang2025pi3}). 
A prevalent paradigm introduces explicit architectural additions to stream these geometric features directly into the MLLM pipeline. VG-LLM~\citep{zheng2025learning(vgllm)} employs direct patch-level feature addition to construct geometry-augmented visual tokens at the input layer. VLM-3R~\citep{fan2025vlm(vlm3r)} concatenates enriched 3D features with camera parameters, subsequently leveraging a cross-attention layer to allow visual tokens to query the geometric context. Meanwhile, GS-Reasoner~\citep{chen2025reasoning(gsreasoner)} incorporates grounding-aware Chain-of-Thought (CoT) supervision to explicitly bridge the gap between object localization and spatial cognition. G$^2$VLM~\citep{hu2025g2vlm} introduces a Mixture-of-Transformer-Experts (MoT) style architecture that bifurcates processing into dedicated geometric and semantic expert modules, optimizing joint 3D reconstruction and high-level reasoning through shared self-attention layers.
Complementary to explicit streaming pipelines, another prominent direction utilizes feature distillation or semantic alignment to bypass the runtime overhead of external models. Frameworks such as 3DRS~\citep{huang20253drs} and 3DThinker~\citep{chen2025think(3dthinker)} distill 3D spatial priors directly from external geometric models into the MLLM's latent visual representations. 
While these distillation methods eliminate the runtime geometry encoder via static supervision, 
they do not expose a separate geometry representation for explicit inference-time routing~\citep{li2026thinking(geothinker)}.
In summary, methods that stream features from external geometry encoders provide explicit structural priors but incur additional runtime computation and latency. 
Training-time distillation avoids this dependency by supervising compact latent representations, but typically does not provide a dense, frame-aligned geometry interface for explicit routing across intermediate layers during inference~\citep{chen2025think(3dthinker)}.
\name bridges this trade-off by using an offline geometry expert for pretraining and generating geometric and temporal priors through internal prediction heads at inference. These self-predicted representations are then selectively integrated through the dual-branch SGT-Fusion module to support task-relevant spatial reasoning over multi-frame visual inputs.

%% file: sections/3_method.tex
\vspace{-3mm}
\section{Method}
\vspace{-1mm}

To eliminate the runtime overhead of external geometry encoding, we propose \name, which introduces Latent Frame Prediction (LFP) heads at intermediate MLLM layers to predict complementary latent representations of geometry and future semantics from causal hidden states. The external geometry encoder is used only as an offline teacher during LFP pretraining and is removed from the subsequent tuning and inference pipelines. In the following, we first present the overall architecture and its decoupled two-stage training strategy, and then describe how the predicted representations are selectively integrated through the Spatial-Grounded Temporal Fusion mechanism.

\vspace{-2mm}
\subsection{Architecture}
\label{sec:method_arch}
\vspace{-1mm}


\textbf{Preliminary.} Given a sequence of RGB images $\left\{I_i\right\}_{i=1}^n$ and a natural-language query $Q$, standard Multimodal Large Language Models (MLLMs) process the input by first projecting pixel-level data into a latent visual space. Specifically, a 2D vision encoder maps each image into semantic visual features $T^S_i\in \mathbb{R}^{\left\lfloor\frac{h}{p_s}\right\rfloor \times  \left\lfloor\frac{w}{p_s}\right\rfloor \times c}$, where $I_i\in \mathbb{R}^{h\times w\times 3}$ and $p_s$ is the patch size. Following recent architectures like Qwen2.5-VL~\citep{Qwen2.5-VL} and Qwen3-VL~\citep{Qwen3-VL}, a spatial compression mechanism is introduced before the LLM layers to enhance computational efficiency. Specifically, given the spatial merge size of $2$, this operation aggregates spatially contiguous $2\times 2$ visual patches into a single representative token, yielding compressed semantic features $T^{S'}_i\in \mathbb{R}^{\left\lfloor\frac{h}{2p_s}\right\rfloor \times  \left\lfloor\frac{w}{2p_s}\right\rfloor \times c}$ to significantly reduce the effective sequence length while preserving local semantic integrity. These tokens are then jointly processed by the LLM layers. Crucially, we denote the intermediate hidden states of the image tokens at a selected LLM layer $j$ as $\textbf{H}_j^{img} \in \mathbb{R}^{n \times L \times c}$ where $L = \lfloor\frac{h}{2p_s}\rfloor \times \lfloor\frac{w}{2p_s}\rfloor$, which serves as the input for our endogenous feature generation.


\textbf{3D Visual Geometry Expert.} To bootstrap our model with structural intuition without explicit 3D supervision, we utilize VGGT~\citep{wang2025vggt} as an offline geometry expert exclusively during the stage 1. Unlike vanilla 2D encoders, this geometry expert captures cross-frame spatial dependencies via a dual-component architecture consisting of an image-wise feature extractor and a cross-frame interaction decoder. Let $p_g$ denote its patch size and it extracts dense geometric feature maps $T_i^{G}\in \mathbb{R}^{\left\lfloor\frac{h}{p_g}\right\rfloor \times  \left\lfloor\frac{w}{p_g}\right\rfloor \times c}$ from all input images jointly. To reconcile the resolution mismatch between the semantic and geometry features, $T_i^{G}$ is resampled and interpolated on a grid aligned with $p_s$ and the $2\times2$ spatial merge factor. This yields a patch-aligned geometric representation $T_i^{G'}\in \mathbb{R}^{\left\lfloor\frac{h}{2p_s}\right\rfloor \times \left\lfloor\frac{w}{2p_s}\right\rfloor \times c}$, ensuring an exact spatial correspondence with the semantic visual grid. 
%
It is worth noting that while $T_i^{G'}$ provides essential structural cues, its serial execution path introduces substantial latency. In our framework, the external encoder is removed at inference and used only offline during Stage 1 to provide geometry-feature targets for training the prediction heads.


\textbf{Latent Frame Prediction.} Inspired by predictive world-modeling paradigms~\citep{yang2025cambrian}, we integrate lightweight Latent Frame Prediction (LFP) MLP heads directly onto the intermediate hidden layers of the MLLM. Instead of relying on an external streaming branch during inference, the model leverages its internal representations to autonomously anticipate spatial geometry and temporal dynamics. Specifically, given the intermediate hidden states $\textbf{H}^{img}_{i,j}$ from the image $I_i$ and selected layer $j$, the dual-targeted LFP heads project the latent features into two distinct latent spaces:
%
\begin{equation}
\hat{T}_i^{G'} = \text{MLP}_{j}^{\text{geo}}(\textbf{H}_j^{img}), \quad \hat{T}^{S'}_{i+1} = \text{MLP}_{j}^{\text{temp}}(\textbf{H}_j^{img}),
\end{equation}
\vspace{-1mm}
where $\hat{T}_i^{G'}$ denotes the predicted representation of current-frame geometry, trained to match the VGGT-derived target $T_i^{G'}$ from \citep{wang2025vggt}, whereas $\hat{T}_{i+1}^{S'}$ denotes the predicted semantic representation of the next frame, trained against the corresponding visual feature target $T_{i+1}^{S'}$.

\input{tables/02_VSI_bench}

\vspace{-2mm}
\subsection{Endogenous Feature Generation}
\vspace{-2mm}

The core insight of \name lies in transitioning from an execution-heavy external geometry branch to an endogenous feature generation pipeline, where the MLLM acts as its own spatial-temporal expert. To achieve this without causing representation drift or disrupting the backbone’s native semantic understanding, we introduce a decoupled two-stage training paradigm:

\textbf{Stage 1: LFP Pre-training.}
As illustrated in Fig.~\ref{fig:model_overview}, Stage 1 aims to cultivate the MLLM's internal geometry and temporal intuition while strictly safeguarding its native multimodal capabilities. To achieve this, we freeze the primary MLLM backbone and exclusively train lightweight LFP heads attached to selected intermediate layers, thereby preventing low-level reconstruction constraints from perturbing top-level token generation throughout the entire pretraining process.

\textbf{Stage 2: Self-Driven Tuning.} 
Transitioning to joint spatiotemporal reasoning, we completely discard the external geometry expert and perform task-specific instruction tuning without external geometry inputs. The geometry and future-semantic representations generated by the LFP heads are routed into the downstream SGT Fusion blocks, enabling the MLLM to solve complex spatial reasoning tasks using entirely endogenous contextual cues within its own latent space.

\input{tables/01_All_Spatial}

\subsection{Spatial Geometry-Temporal Fusion}
Building upon the active perception paradigm of \citep{li2026thinking(geothinker)}, we deploy SGT-Fusion across selected MLLM layers to incorporate the endogenously generated spatial-temporal priors. To effectively handle the distinct characteristics of the synthesized streams, we refine the original fusion mechanism into a dual-modality specialized architecture, which decouples the integration into an Geometry-Prior Branch and a Temporal-Prior Branch. 
Finally, we apply Layer-Interleaved Feature Injection, using learnable layer-wise scaling factors to dynamically calibrate and alternate the integration intensity of these spatial-temporal features.

\subsubsection{Geometry-Prior Branch}
The Geometry-Prior Branch processes endogenously synthesized priors that inherently preserve current-frame spatial geometry, where contiguous feature regions share highly correlated geometric semantics. 
%
More specifically, let $\textbf{H}^{img}_{j}\in \mathbb{R}^{(n\times L) \times c}$ denote the hidden states of the image tokens in the $j$-th layer of LLM, where $L=\left\lfloor\frac{h}{2p_s}\right\rfloor \times  \left\lfloor\frac{w}{2p_s}\right\rfloor$ denotes the token length of each image. Following \citep{li2026thinking(geothinker)}, we predict an importance score $\mathbf{S}_{imp}$ directly from the image hidden states using a lightweight MLP. 
To preserve spatial correspondence and enforce the predictive constraint, we reshape both the image hidden states $\mathbf{H}^{img}_{j}$ and the predicted current-frame spatial geometry features $\{\hat{T}^{G'}_{i}\}_{i=1}^{N}$ into their original spatial dimensions, denoted as $\mathbf{SH}^{img}_{j}$ and $\hat{\mathbf{SG}^{S'}_{j}} \in \mathbb{R}^{n \times L \times c}$, respectively. The query, key, and value projections are then computed such that the current frame's semantics query its own predicted feature, avoiding erroneous queries that span frames. Therefore, the actively perceived geometry features $\mathbf{Geo}_{j}^{img}$ can be computed as:

\vspace{-2mm}
\begin{equation}
    \mathbf{S}^{imp}_{j} = \text{Sigmoid}(\text{MLP}(\mathbf{SH}^{img}_{j})), \\
\end{equation}

\vspace{-2mm}
\begin{equation}
    \!\!\!\!\!\textbf{Q}_{j}^{Geo}\! =\!\! \text{MLP}(\textbf{SH}^{img}_{j}),\
    \textbf{K}_j^{Geo}\! =\!\! \text{MLP}(\hat{\mathbf{SG}_{j}^{S'}}),\!\
    \textbf{V}_j^{Geo}\! =\!\! \text{MLP}(\hat{\mathbf{SG}_{j}^{S'}}).\\
\end{equation}

\vspace{-2mm}
\begin{equation}
   \mathbf{Geo}_{j}^{img} =\! \text{softmax}(\frac{\textbf{Q}^{Geo}_j(\textbf{K}^{Geo}_j)^T}{\sqrt{d_k}}+\log((\mathbf{S}^{imp}_{j})^T+\epsilon))\textbf{V}_j^{Geo}.
\end{equation}

where $d_k$ denotes the dimension of the attention keys.

\subsubsection{Temporal-Prior Branch}

The Temporal-Prior Branch leverages the LFP heads to project and integrate next-frame semantic features, thereby equipping the model with robust cross-frame temporal continuity and forward-looking world-modeling capabilities.
Similarly, we reshape both the image hidden states $\mathbf{H}^{img}_{j}$ and the predicted next-frame temporal dynamics features $\{\hat{T}^{S'}_{i+1}\}_{i=1}^{N}$ into their original spatial dimensions, denoted as $\mathbf{SH}^{img}_{j}$ and $\hat{\mathbf{ST}^{S'}_{j}} \in \mathbb{R}^{n \times L \times c}$, respectively. The query, key, and value projections are then computed such that the current frame's semantics query its own predicted future. Therefore, let $d_k$ denotes the dimension of the attention keys and the constrained cross-attention with temporal dynamic can be formulated as:

\vspace{-1mm}

\begin{equation}
    \!\!\!\!\!\textbf{Q}_{j}^{Tem}\! =\!\! \text{MLP}(\textbf{SH}^{img}_{j}),\
    \textbf{K}_j^{Tem}\! =\!\! \text{MLP}(\hat{\mathbf{ST}_{j}^{S'}}),\!\
    \textbf{V}_j^{Tem}\! =\!\! \text{MLP}(\hat{\mathbf{ST}_{j}^{S'}}).\\
\end{equation}

\vspace{-1mm}

\begin{equation}
   \mathbf{Tem}_{j}^{img} =\! \text{softmax}(\frac{\textbf{Q}^{Tem}_j(\textbf{K}^{Tem}_j)^T}{\sqrt{d_k}})\textbf{V}^{Tem}_j.
\end{equation}

\vspace{-1mm}


\begin{figure*}[t]
	\centering
        \includegraphics[width=1\textwidth]{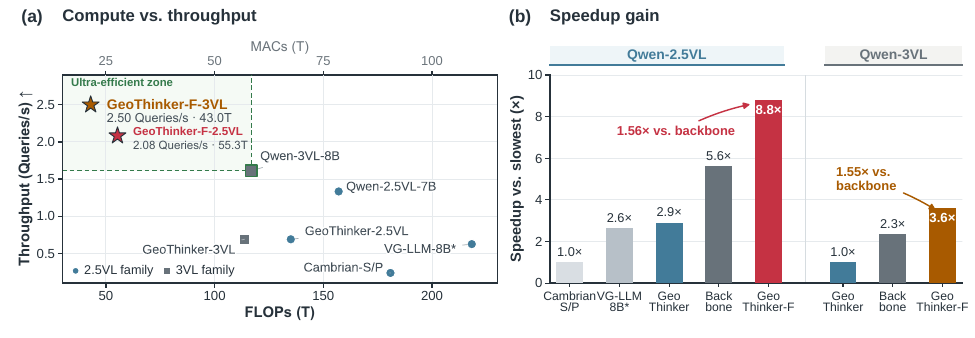}
     \caption{
     \textbf{Efficiency and throughput comparison.} \textbf{(a)} Compute vs. throughput trade-off, showing \name positioned in the ultra-efficient zone with minimal FLOPs and superior throughput. 
     \textbf{(b)} Relative speedup over the highest-accuracy baseline among the compared methods within each backbone family, reaching an $8.8\times$ acceleration on Qwen2.5-VL-7B compared to Cambrian-P and a $3.6\times$ acceleration on Qwen3-VL-8B compared to GeoThinker.
     }
	\label{fig:efficiency_comparison}
    \vspace{-5mm}
\end{figure*}

\vspace{-3mm}
\subsubsection{Layer-Interleaved Feature Injection}
\label{sec:layer_interleaved_feature_injection}
To regulate the intensity of geometric and temporal integration while preventing feature interference, we adopt a layer-interleaved injection strategy managed by a learnable scalar $\alpha_j$ initialized to $0$. Since both geometric and temporal priors are endogenously predicted and inherently noisy compared to feature targets, simultaneous fusion in a single layer leads to error accumulation and representation degradation. To mitigate this noise compound within LFP predicted features, we decouple the multi-modal injection across intermediate transformer layers:

\vspace{-3mm}
\begin{equation}
\hat{\mathbf{H}}_{j}^{\text{img}} = \mathbf{H}_{j}^{\text{img}} + \tanh(\alpha_j) \cdot\begin{cases}\mathbf{Geo}_{j}^{\text{img}}, & \text{if } j \in \mathcal{L}_{\text{geo}} \\ \mathbf{Tem}_{j}^{\text{img}}, & \text{if } j \in \mathcal{L}_{\text{tem}} ~\text{and}~j\notin\mathcal{L}_{\text{geo}}\end{cases}\end{equation}
\vspace{-1mm}
%
where $\mathcal{L}_{\text{geo}}$ and $\mathcal{L}_{\text{tem}}$ represent disjoint layer index sets dedicated to spatial geometry and temporal dynamics respectively, defined by alternating across odd and even LLM layers.
The resulting $\hat{\mathbf{H}}_{j}^{\text{img}}$ serves as the refined hidden state for subsequent multimodal reasoning and response generation.

%% file: tables/02_VSI_bench.tex
\begin{table*}[t]
  \caption{
  \textbf{Performance comparisons on VSI-Bench (Vanilla regime).} By internalizing 3D geometry priors and incorporating temporal dynamics priors, \name completely eliminates the external VGGT encoder while outperforming the GeoThinker.
  }
  \vspace{-3mm}
  \label{tab:vsi-bench-main}
  \begin{center}
    \begin{small}
      \begin{sc}
        \resizebox{\textwidth}{!}{%
        \setlength{\tabcolsep}{1mm}
        \begin{tabular}{l|c|cccc|cccc}
          \toprule
          \multirow{2}{*}{Methods} & \multirow{2}{*}{~Avg.~}
          & \multicolumn{4}{c|}{\cellcolor{orange!10}Numerical Answer}
          & \multicolumn{4}{c}{\cellcolor{yellow!10}Multiple-Choice Answer} \\
          \cmidrule(lr){3-6}\cmidrule(lr){7-10}
          & &  Obj.Count & Abs. Dist. & Obj. Size & Room Size & Rel. Dist. & Rel. Dir. & Route Plan & Appr. Order  \\
          \midrule
          \rowcolor{black!11}
          \multicolumn{10}{l}{\textbf{Baseline}} \\
          Chance Level (Random)    & --   & --   & --   & --   & --   & 25.0   & 36.1   & 28.3   & 25.0 \\
          Chance Level (Frequency)  & 34.0 & 62.1 & 32.0 & 29.9 & 33.1 & 25.1 & 47.9 & 28.4 & 25.2 \\
          \midrule
          \rowcolor{black!8}
          \multicolumn{10}{l}
          {\textbf{Proprietary Models (API)}} \\
          GPT-4o\citep{hurst2024gpt4o}            & 34.0 & 46.2 & 5.3  & 43.8 & 38.2 & 37.0 & 41.3 & 31.5 & 28.5 \\
          Gemini-1.5 Flash\citep{team2024gemini(gemini1.5)}   & 42.1 & 49.8 & 30.8 & 53.5 & 54.4 & 37.7 & 41.0 & 31.5 & 37.8 \\
          Gemini-1.5 Pro\citep{team2024gemini(gemini1.5)}     & 45.4 & 56.2 & 30.9 & 64.1 & 43.6 & 51.3 & 46.3 & 36.0 & 34.6 \\
          \midrule
          \rowcolor{black!11}
          \multicolumn{10}{l}{\textbf{Open-sourced Models}} \\
          LLaVA-OneVision-7B\citep{li2024llava-onevision}       & 32.4 & 47.7 & 20.2 & 47.4 & 12.3 & 42.5 & 35.2 & 29.4 & 24.4 \\
          LLaVA-OneVision-72B\citep{li2024llava-onevision}      & 40.2 & 43.5 & 23.9 & 57.6 & 37.5 & 42.5 & 39.9 & 32.5 & 44.6 \\
          LLaVA-NeXT-Video-7B\citep{zhang2024llavanextvideo}     & 35.6 & 48.5 & 14.0 & 47.8 & 24.2 & 43.5 & 42.4 & 34.0 & 30.6 \\
          LLaVA-NeXT-Video-72B\citep{zhang2024llavanextvideo}   & 40.9 & 48.9 & 22.8 & 57.4 & 35.3 & 42.4 & 36.7 & 35.0 & 48.6 \\
          InternVL2-8B\citep{chen2024internvl2,opengvlab2024internvl2}           & 34.6 & 23.1 & 28.7 & 48.2 & 39.8 & 36.7 & 30.7 & 29.9 & 39.6 \\
          InternVL2-40B\citep{chen2024internvl2,opengvlab2024internvl2}          & 36.0 & 34.9 & 26.9 & 46.5 & 31.8 & 42.1 & 32.2 & 34.0 & 39.6 \\

          
          
          Qwen2.5VL-3B\citep{Qwen2.5-VL}  & 28.6 & 32.7 & 19.5 & 17.3 & 25.1 & 37.3 & 44.9 & 30.4 & 21.8 \\
          Qwen2.5VL-7B\citep{Qwen2.5-VL}  & 29.3 & 25.2 & 10.5 & 36.4 & 29.6 & 38.4 & 38.0 & 29.8 & 26.8 \\
          



          \midrule
          \rowcolor{black!11}
          \multicolumn{10}{l}{\textbf{Open-source Spatial Intelligence Models}} \\


          SPAR-8B\citep{zhang2025flatland(spar)}  & 44.1 & -- & -- & -- & -- & -- & -- & -- & -- \\

          SpatialLadder-3B\citep{li2025spatialladder} & 44.8 & -- & -- & -- & -- & -- & -- & -- & -- \\
          
          Spatial-MLLM-4B\citep{wu2025spatial(spatialmllm)} & 48.4 & 65.3 & 34.8 & 63.1 & 45.1 & 41.3 & 46.2 & 33.5 & 46.3 \\

        
          VG-LLM-4B\citep{zheng2025learning(vgllm)} & 46.7 & 67.6 & 37.6 & 55.2 & 52.5 & 48.0 & 44.7 & 31.9 & 35.5 \\

          VG-LLM-8B~\citep{zheng2025learning(vgllm)}  & 49.7 & 68.1 & 38.7 & 59.0 & 61.1 &45.5 &44.9 &26.8 &53.4\\


          GeoThinker\textsubscript{Qwen2.5VL-7B}~\citep{li2026thinking(geothinker)} & 50.5 & 69.5 & 38.5 & 57.9 & 62.2 & 45.2 & 46.2 & 31.4 & 52.6 \\

        \midrule
          \rowcolor{navyblue!11}
          \multicolumn{10}{l}{\textbf{Ours}} \\
          
          

          

          

          
          \name\textsubscript{Qwen2.5VL-7B}  & \sota{50.7} & 69.8 & 36.2 & 57.6 & 58.1 & 47.9 & 41.8 & 30.9 & 63.6 \\


          \bottomrule
        \end{tabular}%
        }
      \end{sc}
    \end{small}
  \end{center}
  \vspace{-5mm}
\end{table*}

%% file: tables/01_All_Spatial.tex
\begin{table*}[t]
  \caption{\textbf{Cross-benchmark comparison on spatial intelligence benchmarks (Scaled regime).}
  $\dagger$ indicates evaluation on reduced subsets. * indicates trained with official S1+S2 setting. Benchmarks include VSI-Bench\citep{yang2025thinking(vsi)}, MMSI-Bench\citep{yang2026mmsi}, MindCube\citep{yin2025spatial(mindcube)}, Viewspatial\citep{li2025viewspatial}, SITE\citep{wang2025site}, CV-Bench~\citep{tong2024cambrian(cambrian-1)}.
  }
  \vskip -0.1in
  \label{tab:cross-benchmark}
  \begin{center}
    \begin{small}
      \begin{sc}
        \resizebox{\textwidth}{!}{%
        \begin{tabular}{l|c|cccccc}
          \toprule


          \multirow{2}{*}{\textbf{Models}}
          & \multirow{2}{*}{\textbf{~AVG.~}}
          & \multirow{2}{*}{\textbf{VSI-Bench}}
          & \multirow{2}{*}{\textbf{MMSI-Bench}}
          & \multirow{2}{*}{\textbf{MindCube}}
          & \multirow{2}{*}{\textbf{ViewSpatial}}
          & \multirow{2}{*}{\textbf{SITE}}
          & \multirow{2}{*}{\textbf{CV-Bench}}\\

          & & & & & & & \\

          \midrule
          Human & -- & 79.2 & 97.2 & 94.5 & -- & 67.5 & --  \\
          Random Choice & -- & 34.0 & 25.0 & 33.0 & 26.3 & 0.0 & --  \\
          \midrule
          \rowcolor{black!9}
          \multicolumn{8}{l}{\textbf{Proprietary Models}} \\
          Seed-1.6\citep{bytedance2025seed16}
            & 53.41 & 49.9 & 38.3 & 48.7 & 43.8 & 54.6 & 85.2 \\
          Gemini-2.5-Pro\citep{comanici2025gemini25}
            & 56.33 & 53.5 & 38.0 & 57.6 & 46.0 & 57.0 & 85.9 \\
          GPT-5\citep{openai_gpt5_systemcard}
            & 57.50 & 55.0 & 41.8 & 56.3 & 45.5 & 61.8 & 84.6 \\
          Gemini-3-Pro-Preview \citep{gemini_3_pro_systemcard}
            & 62.16 & 52.5 & 45.2 & 70.8 & 50.3 & 62.2 & 92.0 \\
          \midrule
          \rowcolor{black!9}
          \multicolumn{8}{l}{\textbf{Open-source General Models}} \\
          Bagel-7B-MoT\citep{deng2025bagel}
            & 41.90 & 31.4 & 31.0 & 34.7 & 41.3 & 37.0 & 76.0\\
          Qwen2.5-VL-3B-Instruct\citep{Qwen2.5-VL}
            & 38.60 & 28.6 & 28.6 & 37.6 & 31.9 & 33.1 & 71.8 \\
          Qwen2.5-VL-7B-Instruct\citep{Qwen2.5-VL}
            & 40.31 & 29.3 & 26.8 & 36.0 & 36.8 & 37.6 & 75.4  \\
          Qwen3-VL-2B-Instruct\citep{Qwen3-VL}
            & 42.40 & 49.4 & 11.9 & 31.4 & 34.2 & 35.6 & 78.4 \\
          Qwen3-VL-8B-Instruct\citep{Qwen3-VL}
            & 47.70 & 57.7 & 28.8 & 29.8 & 39.0 & 45.8 & 85.1 \\
          InternVL3-2B\citep{zhu2025internvl3}
            & 39.31 & 32.9 & 26.5 & 37.5 & 32.5 & 30.0 & 76.5 \\
          InternVL3-8B \citep{zhu2025internvl3}
            & 45.38 & 42.1 & 28.0 & 41.5 & 38.6 & 41.1 & 81.0 \\
          \midrule
          \rowcolor{black!9}
          \multicolumn{8}{l}{\textbf{Open-source Spatial Intelligence Models}} \\
          SpatialLadder-3B\citep{li2025spatialladder}
            & 42.83 & 44.8 & 27.4 & 43.4 & 39.8 & 27.9 & 73.7 \\
          VST-3B-SFT \citep{vst2025}
            & 49.50 & 57.9$\dagger$ & 30.2$\dagger$ & 35.9 & 52.8 & 35.8 & 84.4\\
          VST-7B-SFT\citep{vst2025}
           & 51.31 & 60.6$\dagger$ & 32.0$\dagger$ & 39.7 & 50.5 & 39.6 & 85.5   \\
          Cambrian-S-3B\citep{yang2025cambrian}
            & 42.91 & 57.3 & 25.2 & 32.5 & 39.0 & 28.3 & 75.2 \\
          Cambrian-S-7B \citep{yang2025cambrian}
            & 47.28 & 67.5 & 25.8 & 39.6 & 40.9 & 33.0 & 76.9 \\

            Cambrian-P \citep{yang2026cambrian(cambrianP)}
            & - & 73.7 & 28.0 & 38.4 & - & - & - \\

            SenseNova-SI-1.1\textsubscript{Qwen3VL-8B}~\citep{cai2025scaling(sensenova)} & - & 64.8 & 38.1 & 73.8 & 51.2 & 49.6 & - \\
          SenseNova-SI-1.1\textsubscript{InternVL3-8B}~\citep{cai2025scaling(sensenova)} & - & 68.8 & 43.3 & 85.7 & 54.7 & 47.7 & - \\

          3DThinker\textsubscript{Qwen2.5-7B}\citep{chen2025think(3dthinker)} & - & 63.7 & 43.3 & 76.0 & 68.6  & -  & 81.1 \\

            
          VLM-3R-7B\citep{fan2025vlm(vlm3r)}
           & 45.40 & 60.9 & 27.9 & 40.0 & 40.5 & 31.3 & 71.8 \\
          VG-LLM-4B\citep{zheng2025learning(vgllm)} & 47.46 & 46.6 & 28.0 & 36.9 & 42.5  & 49.8  & 81.0\\
          VG-LLM-8B\citep{zheng2025learning(vgllm)} & 48.15 & 49.6 & 28.4  & 32.7  & 42.9  & 52.6  & 82.7  \\
          VG-LLM-8B*\citep{zheng2025learning(vgllm)} & 51.05 & 62.2 & 30.0 & 36.1 & 45.8 & 50.5 & 81.7 \\
          
            

            GeoThinker\textsubscript{Qwen2.5VL-7B}~\citep{li2026thinking(geothinker)} & 60.43 & 68.5 & 31.7 & 83.6 & 41.9 & 54.8 & 82.1 \\
            
            GeoThinker\textsubscript{Qwen3VL-8B}~\citep{li2026thinking(geothinker)} & 62.23 & 72.6 & 30.9  & 83.0  & 45.9 & 55.9 & 85.1  \\

            \midrule
            \rowcolor{navyblue!9}
            \multicolumn{8}{l}{\textbf{Ours}} \\
            \name\textsubscript{Qwen2.5VL-7B} & \sota{66.84} & 67.3 & 31.2 & 86.8 & 83.2 & 53.2 & 79.1 \\
            
            \name\textsubscript{Qwen3VL-8B} & \sota{69.29} & 70.4 & 30.8  & 93.2  & 82.6 & 53.7 & 85.0  \\

          \bottomrule
        \end{tabular}}
      \end{sc}
    \end{small}
  \end{center}
  \vspace{-5mm}
\end{table*}

%% file: sections/4_experiment.tex
\begin{figure*}[t!]
     \centering

     \includegraphics[width=\textwidth]{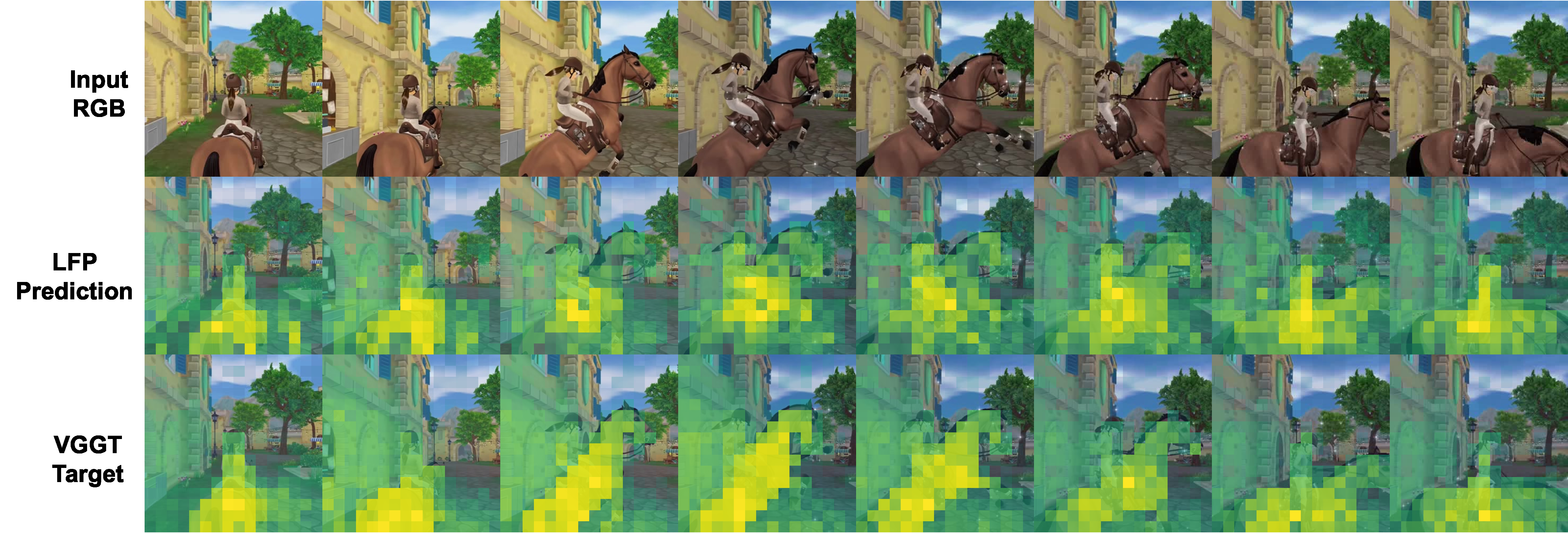}
     \caption{
     \textbf{Qualitative comparison between predicted geometry representations and VGGT feature targets.} From top to bottom, we show the input RGB frames, geometry representations predicted by LFP head, and the corresponding feature targets from offline VGGT expert. The predicted features demonstrate remarkable structural alignment with the offline-expert feature targets.
     }
	\label{fig:visualization}
    \vspace{-5mm}
\end{figure*}

\vspace{-2mm}
\section{Experiments}
\vspace{-1mm}
In this section, we first provide implementation details in \ref{sec:implement_details}, followed by evaluation results on spatial reasoning benchmarks in \ref{sec:evaluation_results}, demonstrating the effectiveness of our approach. 
Next, we conduct an ablation study in \ref{exp:ablation} to verify the contribution of each component in \name. Finally, in \ref{exp:visualization}, we provide visualizations of our core designs to better interpret model behavior.


\vspace{-2mm}
\subsection{Implementation Details.}
\label{sec:implement_details}
\vspace{-1mm}
\textbf{Baseline.} 
GeoThinker~\citep{li2026thinking(geothinker)} propose active geometry integration to actively select geometry features from VGGT~\citep{wang2025vggt} as demand for MLLMs, serving as our primary baseline.
\textbf{(1) Vanilla regime:} Following VG-LLM configuration, we utilize sampled subsets from SPAR-7M \citep{zhang2025flatland(spar)} and the LLaVA-Hound split of LLaVA-Video-178K \citep{zhang2024video} for fine-tuning. We uniformly sample 8 frames per scene for consistency with the baseline.
\textbf{(2) Scaled regime:} To probe the performance ceiling, we scale the training set with data from VLM-3R \citep{fan2025vlm(vlm3r)}, VSI-590K \citep{yang2025cambrian}, MindCube \citep{yin2025spatial(mindcube)} and OmniSpatial\citep{jia2026omnispatial}. 
We increase the sampling density to 32 frames per scene, and additionally incorporate 430k general video samples from~\citep{yang2025cambrian} to strengthen video understanding.

\textbf{Implementation Details.} 
To isolate the effects of our method and pre-training data, all experiments use the same subset of LLaVA-Hound in Vanilla regime for Stage 1 pretraining, ensuring that performance differences are not attributable to unequal amounts of instruction data introduced before SFT.
To better assess the effectiveness and generalization of our design, we test \name across multiple VLM backbones under different training regimes~\citep{li2026thinking(geothinker)}. For spatial reasoning, we adopt three incremental training settings, all using a batch size of 64 and a learning rate of 1e-5. First, following VG-LLM \citep{zheng2025learning(vgllm)}, the training step is set to 4,656. Next, by scaling up the VSI-Bench instruction data, we increase the training steps to 19,315. Finally, we further incorporate general video data from \citep{yang2025cambrian}, which brings the total training steps to 28,235. 
The fusion ratio $\rho$ is set to 1 .
%
%
All experiments are conducted on 16 PPU-ZW810E 96GB PPUs.

\vspace{-2mm}
\subsection{Evaluation Results}
\label{sec:evaluation_results}
\vspace{-1mm}

We conduct evaluation across multiple benchmarks, including VSI-Bench~\citep{yang2025thinking(vsi)}, MMSI-Bench~\citep{yang2026mmsi}, MindCube~\citep{yin2025spatial(mindcube)}, ViewSpatial~\citep{li2025viewspatial}, SITE~\citep{wang2025site}, and CV-Bench~\citep{tong2024cambrian(cambrian-1)}.

\textbf{Vanilla regime:} To evaluate the effectiveness of our endogenous approach under a controlled setup, we conduct experiments on VSI-Bench following the evaluation protocols established in~\citep{zheng2025learning(vgllm),li2026thinking(geothinker)}. For a fair comparison, we retain the identical VLM backbone and training data split, substituting external geometry feature streaming with our endogenous perception architecture. 
As reported in Table~\ref{tab:vsi-bench-main}, \name achieves an average score of \textbf{50.7}, outperforming both the input-level fusion baseline VG-LLM-8B at 49.7 and the dual-encoder GeoThinker baseline at 50.5. Crucially, \name achieves these top-tier results without requiring an external geometry encoder at runtime. 
This performance gain confirms that endogenously synthesized geometric textures and dynamic temporal priors, when dynamically routed through our SGT-Fusion module, supply sufficient spatial-temporal intuition while bypassing external multi-encoder overhead.

\textbf{Scaled regime:} To probe the scalability and generalization of our architecture, we expand the training mixture by integrating scaled spatial instruction data alongside general video samples. To verify that \name acquires versatile spatial intelligence rather than benchmark-specific overfitting, we evaluate across a comprehensive suite of six spatial reasoning benchmarks and prioritize cross-benchmark average accuracy. As shown in Table~\ref{tab:cross-benchmark}, \name sets a new state-of-the-art among open-source spatial models, outperforming both explicit-encoder baselines and leading proprietary API models (e.g., Gemini-3-Pro-Preview at $62.16$). Specifically, \name\textsubscript{Qwen2.5VL-7B} and \name\textsubscript{Qwen3VL-8B} achieve peak average scores of \textbf{66.84} and \textbf{69.29}, respectively. 
Notably, our model exhibits dramatic gains on complex 3D mental modeling and multi-perspective tasks that reaches $93.2$ on MindCube and $82.6$ on ViewSpatial, confirming that endogenously synthesized spatial-temporal priors provide robust, transferable spatial representations across diverse scenarios.

\vspace{-1mm}

\vspace{-1mm}
\subsection{Ablation study}
\label{exp:ablation}
\input{tables/06_ablation_study}
\vspace{-1mm}

To evaluate the effects of LFP pretraining and the two predicted priors, we conduct a controlled ablation study on VSI-Bench using Qwen2.5-VL-7B. All variants are trained with the same vanilla SFT data. As reported in Table~\ref{tab:s1_ablation}, the SFT baseline without LFP heads or SGT-Fusion achieves an average accuracy of $49.36\,\%$. Adding both prediction heads and the fusion modules without Stage-1 LFP pretraining yields a comparable score of $49.32\,\%$, indicating that the improvement does not arise merely from introducing additional trainable parameters. With LFP pretraining, the Geometry Branch and Temporal Branch individually achieve $50.17\,\%$ and $50.54\,\%$, respectively. Combining both branches produces the best average accuracy of \textbf{$50.76\,\%$}. These results validate the importance of Stage-1 LFP pretraining and the effectiveness of both predicted representations for spatial reasoning within a controlled and consistent SFT training configuration.

\vspace{-2mm}
\subsection{Visualization}
\label{exp:visualization}
\vspace{-1mm}

To qualitatively examine whether the LFP heads learn geometry-related structure from the offline expert, we visualize the predicted geometry representations alongside their corresponding VGGT feature targets~\citep{wang2025vggt}. As illustrated in Figure~\ref{fig:visualization}, the predicted representations consistently localize the dominant foreground structure and follow its evolving position and spatial extent across consecutive frames. Their frame-wise activation layouts correspond closely with the VGGT targets, particularly in capturing the horse-rider foreground and its structural changes throughout the sequence. These visualizations suggest that the LFP heads effectively learn the principal spatial organization and frame-wise structural regularities encoded by the VGGT features.

%% file: tables/06_ablation_study.tex
\begin{table}[t] 
    \captionsetup{type=table}
    \centering 
    \caption{
    \textbf{Ablation Study on VSI-Bench.} 
    All compared variants are trained on the same vanilla SFT data with Qwen2.5-VL-7B backbone, showing the individual and joint contributions of our endogenous Geometry-Prior LFP and Temporal-Prior LFP branches.
    }
    \begin{minipage}{0.98\textwidth} 
    \centering 
    \resizebox{\linewidth}{!}{%
        \fontsize{4.6pt}{4.4pt}\selectfont
        \setlength\tabcolsep{3pt}%
        \renewcommand{\arraystretch}{1.4}%
        \begin{tabular}{ccc|c|cccccccc}

         & & & & 
        \rotatebox{45}{Obj. Count} &
        \rotatebox{45}{Abs. Dist.} &
        \rotatebox{45}{Obj. Size} &
        \rotatebox{45}{Room Size} &
        \rotatebox{45}{Rel. Dist.} &
        \rotatebox{45}{Rel. Dir.} &
        \rotatebox{45}{Route Plan} &
        \rotatebox{45}{Appr. Order} \\

        
        \textbf{LFP Pretrain} & \textbf{Geometry} & \textbf{Temporal} & Avg. &
        \multicolumn{4}{c}{\cellcolor{orange!10}Numerical Answer} &
        \multicolumn{4}{c}{\cellcolor{yellow!10}Multiple-Choice Answer} \\
        \hline
        


        \xmarkg & \xmarkg & \xmarkg  & 49.36 & 67.6 & 36.2 & 58.9 & 60.8 & 42.6 & 45.6 & 26.2 & 56.6 \\

        \xmarkg & \cmarkg & \cmarkg & 49.32 & 69.4 & 36.8 & 57.7 & 60.5 & 46.7 & 35.6 & 30.4 & 57.1 \\

        
        
        \cmark & \cmark & \xmarkg  & 50.17 & 70.0 & 35.8 & 57.0 & 58.9 & 47.1 & 41.5 & 32.5 & 58.4 \\
        
        \cmark & \xmarkg & \cmark  & 50.54 & 70.1 & 37.1 & 57.1 & 59.9 & 46.6 & 41.8 & 31.9 & 59.7 \\
        
        \rowcolor{navyblue!10} 
        \cmark & \cmark & \cmark  & \sota{50.76} & 69.8 & 36.2 & 57.6 & 58.1 & 47.9 & 41.8 & 30.9 & 63.6 \\
        \hline
        \end{tabular}%
        }  
    \end{minipage}
    \label{tab:s1_ablation}
    \vspace{-3mm}
\end{table}

%% file: sections/5_conclusion.tex
\vspace{-2mm}
\section{Conclusion}
\vspace{-1mm}

In this paper, we introduced \name, an efficient framework for geometry-aware spatial reasoning that replaces runtime external geometry extraction with endogenous latent prediction. Through a decoupled two-stage training strategy, Latent Frame Prediction (LFP) heads learn from an offline geometry expert and next-frame semantic targets to predict complementary representations of current geometry and future semantics from intermediate MLLM states. These self-predicted representations are selectively integrated through the Geometry-Prior and Temporal-Prior Branches of the Spatial-Grounded Temporal (SGT) Fusion module. 
\name achieves performance comparable to GeoThinker without executing its external geometry encoder at inference. Across six spatial intelligence benchmarks, \name reaches an average score of $69.29$. When combined with spatial token compression, the complete system reduces inference latency from $1.45\,\text{s}$ to $0.40\,\text{s}$ on Qwen3-VL-8B, corresponding to a $3.6\times$ speedup. These results demonstrate that external geometry models can serve as training-time teachers, while useful geometric and temporal representations are generated internally with substantially lower runtime overhead.

\newpage
\clearpage

\subsection*{AI Use Statement}

In this work, we used generative AI tools to provide feedback on experimental design, assist with the implementation and editing of research code, interpret experimental results, and support grammar and language editing. We used these tools to draft and edit portions of the manuscript, improve readability, propose alternative titles and organizational structures, and brainstorm high-level concepts for scientific figures. 
AI tools also assisted with literature searches and bibliographic verification. All references were reviewed by the authors.
The tools were not used to originate the core research hypothesis or scientific findings, design the proposed model or training strategy, generate or clean datasets, create synthetic data, formulate mathematical claims or proofs. 
Generative AI tools assisted with the design and initial creation of selected scientific figures. Most figures were manually created, and all AI-assisted figures were carefully reviewed, substantially edited, and finalized by the authors.
All AI-assisted text and code were reviewed by the authors, the code was tested for correctness, and technical descriptions, experimental interpretations, and numerical claims were checked against the implementation and experimental results. We take responsibility for the final content of this work, including text, claims, code, and artifacts produced with the aid of generative AI.

\subsection*{Ethics Statement}

This work uses publicly available research datasets and pretrained models and does not involve new data collection or interaction with human subjects. We use these resources in accordance with their respective licenses and terms of use. The qualitative visualizations are derived exclusively from video samples released as part of LLaVA-Hound and are presented in accordance with its documented conditions for research use. We do not collect additional personal information, attempt to identify depicted individuals, or infer sensitive attributes from their appearance. The proposed method is evaluated as a research system for spatial reasoning and is not intended for direct deployment in safety-critical environments without additional validation. As with other multimodal models, its behavior may inherit limitations or biases present in the underlying pretrained models and public datasets. We are not aware of other specific ethical concerns arising from this work.

\subsection*{Reproducibility Statement}

The main paper describes the overall architecture, two-stage training procedure, and evaluation settings. Additional implementation details, including model configurations, data composition, training objectives, computational costs, and efficiency analyses, are provided in the appendix. All experiments use publicly available datasets and a fixed random seed. For the reported efficiency results, latency is measured on a PPU-ZW810E 96GB accelerator using FP16 precision, a batch size of one, and a fixed 32-frame input in the spatial reasoning setting. We perform 10 warm-up runs followed by 100 measured runs and report the average latency. The corresponding input-token configurations are documented in the appendix. The source code, trained models, configurations, and data-processing resources will be made publicly available to facilitate reproduction of the reported results. The code and models will be open-sourced upon acceptance.

%% file: sections/6_appendix.tex
\appendix
\clearpage
\newpage

\section{Appendix/supplemental material}
The outline of the Appendix is as follows:
\begin{itemize}
    \item More implementation details;
    \item More analysis on downstream scenarios validation;
    \begin{itemize}
        \item Additional analysis on autonomous driving;
        \item Additional analysis on embodied manipulation;
    \end{itemize}
    \item More analysis on computational cost;
    \item More analysis on general video mixture and test-frame scaling;
    \begin{itemize}
        \item Additional analysis on general video mixture;
        \item Additional analysis on test-frame scaling;
    \end{itemize}
    \item More evaluation on VSTI-Bench;
    \item More evaluation on EASI Leaderboard;
    \begin{itemize}
        \item More evaluation on EASI-8;
        \item More evaluation on EASI-All;
    \end{itemize}
    \item More evaluation on ReVSI-Bench;
    \item More ablation study on OmniSpatial Training data;
    \item More ablation study on Stage-2 training;
    \item More analysis on Stage 1 Pre-training Loss Function;
    \begin{itemize}
        \item Additional definition on basic operations;
        \item Additional definition on local relation loss;
        \item Additional analysis on pre-training loss of temporal branch;
        \item Additional analysis on pre-training loss of geometry branch;
    \end{itemize}
    \item More visualization comparisons;
    \item More discussion;
        
\end{itemize}

\section{Implementation Details}
\label{sec:appendix_implementation_details}

\subsection{Model Configurations}

We conduct all evaluations based on the 32-frame training setting using \name\textsubscript{Qwen25VL-7B} and \name\textsubscript{Qwen3VL-8B} in scaled regime. Below, we elaborate on the specific architectural configurations, feature injection protocols, and optimization strategies deployed across different domain regimes:

\textbf{Layer-Wide Interleaved Feature Injection}:
Unlike prior designs that adjust the injection depth according to the training-data scale~\citep{li2026thinking(geothinker)}, \name applies Layer-Interleaved Feature Injection across all ($\rho=1.0$) intermediate Transformer layers. Importantly, full-depth coverage does not expose every layer to geometric features. Instead, current-geometry and future-semantic predictions are injected into alternating layers, limiting direct geometry injection to approximately half of the network depth. Because the temporal branch predicts next-frame semantic representations aligned with the MLLM's visual-semantic space, it provides a semantically compatible complement to the geometry branch. We further assign the final injection layer to the temporal branch, keeping the uppermost hidden states semantically anchored before language decoding. This interleaved design enables consistent spatiotemporal guidance throughout the backbone while reducing potential interference from dense geometry injection.

\textbf{Domain-Adaptive LFP Head Optimization}: 
To balance cross-domain generalization and in-domain adaptability, the optimization status of the Latent Frame Prediction (LFP) heads is dynamically configured:
\begin{itemize}
    \item \textit{Out-of-Domain (OOD) Transfer}: The LFP heads remain \textbf{frozen} during downstream tuning. 
    This preserves the geometry and temporal representations acquired during pretraining and reduces their drift during downstream tuning.
    \item \textit{In-Domain (ID) Adaptation}: The LFP heads are \textbf{unfrozen} and end-to-end fine-tuned along with the fusion layers, allowing the heads to adaptively align with domain-specific geometric and motion distributions.
\end{itemize}

\textbf{Adaptive Spatial Token Compression}: 
To efficiently scale to long temporal horizons (e.g., 32 frames per scene) without exceeding memory boundaries, we incorporate an adaptive spatial token compression strategy following \citep{li2026thinking(geothinker)}. Rather than statically compressing all inputs, we introduce a sequence-length condition:
\begin{itemize}
    \item \textit{Short Sequences ($\le 8$ frames)}: A heuristic bypass is triggered to skip spatial compression entirely, safeguarding fine-grained visual and geometric details for sparse frame inputs.
    \item \textit{Long Sequences ($> 8$ frames)}: Spatial compression is activated in conjunction with Importance Gating mechanism. By dynamically filtering redundant visual tokens, IG allows the spatial merge stride to increase from 2 to 4 while retaining geometry-relevant information.
\end{itemize}

\subsection{Data Curation}

For the In-Domain training of our model, we curated a large-scale multimodal dataset totaling 1.7M samples. The data composition is as follows:

\textbf{Spatial Reasoning:} Cambrian-S VSI-bench instruction (590k), SPAR (234k), VLM-3R VSI-bench instruction (205k), VLM-3R VSTI-bench instruction (132k), MindCube training set (10k) and OmniSpatial training set (7k).

\textbf{General Video:} LLaVA-Hound (64k) and a subset of general video data sampled from Cambrian-S-3M (430k).




\section{Downstream Scenarios Validation}
\label{exp:downstream}

\vspace{-1mm}
\subsection{Autonomous Driving}

\input{tables/05_recogdrive}



\textbf{Baseline.} ReCogDrive~\citep{li2025recogdrive} is a cognitive framework designed for end-to-end autonomous driving. 
In our implementation, we focus on its spatial planning capabilities during VLM pre-training without incorporating subsequent diffusion planners or reinforcement learning pipelines. 
For evaluation, we conduct non-reactive simulation experiments on NAVSIM \textit{navtest}~\citep{dauner2024navsim} to assess safety-critical decision-making intelligence.

\noindent \textbf{Results.} As reported in Table~\ref{tab:comparison_recogdrive}, \name significantly mitigates prior integration overhead while enhancing downstream driving safety. Specifically, \name reduces the auxiliary latency overhead introduced by spatial prior fusion by \textbf{52.91\%} from $42.62\,\text{ms}$ to $37.08\,\text{ms}$ and lowers total computational cost from $122.69\,\text{T}$ to $88.22\,\text{T}$ FLOPs. 
Notably, even without deploying spatial token compression, which serves as an orthogonal strategy for further inference acceleration in our implementation in \ref{sec:implement_details}, \name natively achieves a $>50\%$ reduction in integration overhead by completely bypassing external encoder execution.
In terms of decision-making quality, \name improves NC, DAC, TTC, and PDMS while matching the baseline on Comfort and EP, yielding higher Drivable Area Compliance (DAC: $94.0\%$ vs. $93.7\%$) and Time-To-Collision compliance (TTC: $90.4\%$ vs. $87.8\%$), elevating the overall PDMS score to $83.3\%$. These results confirm that endogenously synthesized spatial-temporal priors provide a lightweight, lower-overhead alternative for autonomous planning.

\subsection{Embodied Manipulation}
We evaluate embodied manipulation on LIBERO~\citep{liu2023libero} using the StarVLA framework~\citep{community2026starvla}.
\input{tables/20_libero}

\noindent \textbf{Results.} As reported in Table~\ref{tab:apendix_libero_results}, \name achieves strong task-execution performance while substantially reducing the inference overhead of explicit spatial-prior integration. Relative to the $82.0\,\text{ms}$ StarVLA-OFT backbone, GeoThinker introduces $172.8\,\text{ms}$ of additional latency, whereas \name introduces only $22.2\,\text{ms}$. \name therefore reduces the spatial-prior integration overhead by \textbf{87.15\%}, lowering end-to-end inference latency from $254.8\,\text{ms}$ to $104.2\,\text{ms}$ and computational cost from $5.84\,\text{T}$ to $2.51\,\text{T}$ FLOPs, close to the $2.44\,\text{T}$ FLOPs of the raw StarVLA-OFT backbone. Across the four LIBERO task suites, \name attains a $97.1\%$ average success rate, with $97.6\%$ on Spatial, $99.6\%$ on Object, $98.2\%$ on Goal, and $93.0\%$ on Long. It surpasses StarVLA-OFT ($96.6\%$) and GeoThinker ($96.7\%$), while matching OpenVLA-OFT-7B ($97.1\%$) with over $5\times$ fewer training steps ($30\text{K}$ versus $175\text{K}$). These results show that endogenous geometry and temporal representations can retain strong embodied task performance with substantially lower inference overhead.

\section{Computational cost}

\subsection{Spatial Intelligence}
\input{tables/09_appendix_flops}

To elucidate the efficiency bottlenecks inherent to 3D spatial intelligence, we analyze the computational and latency profiles across different model architectures on multi-view benchmarks like VSI-Bench in Table~\ref{tab:appendix_flops}. For GeoThinker and \name, \emph{w/o further token compression} retains the base long-video tokenization, while the complete variants apply the additional compression strategy described in Appendix ~\ref{sec:appendix_implementation_details}. Other models are evaluated using their native tokenization settings. Our causal efficiency comparison is therefore made primarily within matched backbone and model-family configurations.

\paragraph{Input-Output Asymmetry and the Prefill Bottleneck.}
Achieving a holistic 3D understanding of an unconstrained scene necessitates evaluating dense multi-view video streams to avoid missing critical spatial cues. 
Following the standard evaluation protocols in VG-LLM, processing a representative 32-frame video sequence entails a substantial input context of $8,192$ tokens, which comprises $8,064$ visual tokens across the 32 frames and $128$ text prompt tokens.
However, the target answers for benchmark queries are often concise single-token predictions (e.g., multiple-choice option letters or single numerical values). Under this extreme input-output asymmetry, autoregressive decoding latency becomes practically negligible, and total execution time is overwhelmingly dominated by visual encoding and prior extraction during the prefill phase.

\paragraph{Impact of Image Count vs. Sequence Length.}
A key empirical insight from comparing multi-frame baselines in Cambrian-P variants is that the number of input visual frames exerts a far more severe latency penalty than raw token sequence length alone. Processing a large frame count forces standard vision encoders or explicit geometry models (e.g., VGGT) to execute repeated high-resolution forward passes, creating a massive pre-LLM bottleneck. 
Latent-token distillation models such as 3DThinker eliminate runtime 3D encoders. Under the optimistic lower-bound convention described in the caption, we assign 3DThinker the latency of its unaugmented backbone ($0.60\,\text{s}$ for Qwen2.5-VL-3B and $0.75\,\text{s}$ for Qwen2.5-VL-7B), excluding potential output-length-dependent overhead from latent-token decoding. However, these compact latent representations do not provide a dense, frame-aligned geometry interface for explicit routing across intermediate layers.
Conversely, explicit dual-encoder frameworks like GeoThinker suffer severe latency spikes ($1.45\,\text{s}$ to $1.69\,\text{s}$), limiting deployment in latency-sensitive settings.

\paragraph{Synergistic Acceleration via Endogenous Priors and Token Merging.}
Recognizing that external geometry encoders create an unsustainable prefill bottleneck and that dense multi-image streams harbor substantial spatial redundancy, \name synergizes endogenous LFP feature generation with spatial token compression. As shown in Table~\ref{tab:appendix_flops}, on the Qwen3VL-8B backbone with 32-frame setting, purely eliminating the external VGGT encoder slashes latency from $1.66\,\text{s}$ down to $0.91\,\text{s}$ at \name\textsubscript{Qwen3VL-8B-32frame} w/o token merge. When combined with spatial token merging, \name\textsubscript{Qwen3VL-8B-32frame} reduces total FLOPs from $113.77\,\text{T}$ to just $43.01\,\text{T}$ and drives end-to-end latency down to $\mathbf{0.40\,\text{s}}$, delivering a $\mathbf{3.6\times}$ speedup over GeoThinker ($1.45~\text{s}\rightarrow0.40~\text{s}$) and lower latency than the unaugmented Qwen3-VL-8B backbone ($0.62~\text{s}$), while retaining a competitive VSI score of $70.4\%$.

\subsection{Autonomous Driving}
\input{tables/21_appendix_recogdrive_latency}

To provide deeper insight into the computational efficiency of \name in downstream applications, we present a detailed breakdown of parameter counts, FLOPs, and latency for autonomous driving planning on the NAVSIM benchmark in Table~\ref{tab:appendix_recogdrive_latency}.

\paragraph{Experimental and Token Configuration.}
All models are evaluated under a standardized setup matching training configurations, utilizing 16 dynamic visual patches. 
In a representative benchmark scenario, the model processes an input context of $3,272$ tokens comprising $3,072$ visual tokens and $200$ text tokens and autoregressively generates $157$ planning trajectory tokens.
Notably, due to the requirement for fine-grained spatial control in safety-critical autonomous driving, spatial token compression was deliberately omitted in this downstream setup. Consequently, visual tokens are retained at full density, presenting a stringent stress test for model efficiency.

\paragraph{Parameter and FLOPs Reduction.}
As detailed in Table~\ref{tab:appendix_recogdrive_latency}, the explicit dual-encoder baseline, GeoThinker, requires maintaining a heavy runtime external VGGT encoder, contributing an additional $2,076.1\,\text{M}$ parameters and $49.23\,\text{T}$ FLOPs beyond the base MLLM backbone. In contrast, \name replaces explicit 3D feature streaming with lightweight Latent Frame Prediction (LFP) heads and SGT-Fusion modules. This reduces auxiliary parameters to $1,187.1\,\text{M}$ and slashes auxiliary FLOPs to just $14.77\,\text{T}$, achieving a \textbf{69.99\% reduction in additional computational cost} compared to GeoThinker ($49.23\,\text{T} \rightarrow 14.77\,\text{T}$).

\paragraph{Latency Breakdown and Amortization Dynamics.}
In terms of runtime latency, GeoThinker incurs a significant extra execution delay of $1,461.42\,\text{ms}$ due to serial multi-encoder pre-processing. By eliminating external encoder execution, \name reduces this auxiliary latency to $598.20\,\text{ms}$, achieving a \textbf{59.07\% reduction in additional prior integration time}. 

When examining the amortized end-to-end latency per generated planning token (e.g., $32.15\,\text{ms}$ for ReCogDrive, $42.62\,\text{ms}$ for GeoThinker, and $37.08\,\text{ms}$ for \name), the relative gap narrows slightly. This behavior occurs because the autoregressive decoding phase generates $157$ consecutive tokens. Thus, the fixed one-time visual encoding and prior synthesis latency is amortized across the entire decoding sequence. Nevertheless, in real-time robotic and autonomous driving systems where perception and pre-processing throughput dictate latency boundaries, the absolute reduction of over $869.60\,\text{ms}$ in total frame execution ($6,690.67\,\text{ms} \rightarrow 5,821.07\,\text{ms}$) can improve responsiveness in latency-sensitive deployment settings.

\input{tables/22_appendix_libero_latency}
\begin{minipage}[h]{\textwidth}

\begin{minipage}[h]{0.58\textwidth}
\makeatletter\def\@captype{table}
   \setlength{\tabcolsep}{0.1mm}
\caption{
\textbf{Impact of general video data co-training.} We evaluate performance changes ($\Delta$) when co-training with 430k general video instruction samples. While standard 2D MLLM baselines exhibit task interference of degrading 3D spatial accuracy on VSI, \name effectively mitigates cross-task degradation. By internalizing spatial-temporal priors directly within hidden representations, \name maintains robust spatial intelligence while simultaneously enhancing general temporal understanding.
}
\resizebox{1\linewidth}{!}{
\begin{tabular}{lcccc}
\toprule
\textbf{Model} & \textbf{Video Mixture} & \textbf{VSI} & \textbf{VideoMME} & \textbf{MVBench} \\

\midrule
\multirow{2}{*}{Cambrian-S-7B}
& \xmark & 69.2 & 54.1 & - \\
& 3M & 65.1\textcolor{orange}{\textbf{(-4.1)}} & 61.9\textcolor{blue}{\textbf{(+7.8)}} & 64.5 \\

\midrule
\multirow{2}{*}{GeoThinker\textsubscript{Qwen3vl-8B-32frame}}
& \xmark & 72.0 & 53.7 & 42.8 \\
& 430k & 72.6\textcolor{blue}{\textbf{(+0.6)}} & 59.4\textcolor{blue}{\textbf{(+5.7)}} & 69.1\textcolor{blue}{\textbf{(+26.3)}} \\

\midrule
\multirow{2}{*}{\name\textsubscript{Qwen3vl-8B}}
& \xmark & 70.00 & 55.9 & 55.8 \\
& 430k & 70.4\textcolor{blue}{\textbf{(+0.4)}} & 58.7\textcolor{blue}{\textbf{(+2.8)}} & 56.9\textcolor{blue}{\textbf{(+1.1)}} \\

\bottomrule
\end{tabular}}
  \label{tab:ablation-study-structure}
\end{minipage}
    \hspace{8pt}
\begin{minipage}[h]{0.42\textwidth}
\makeatletter\def\@captype{table}
 \addtolength{\tabcolsep}{-3pt}
 \caption{
\textbf{Inference-time frame-count scaling on VSI and VSI-Debiased.} 
Models trained on up to $32$-frame windows are evaluated under scaling inference frames up to 128 frames to assess temporal context extrapolation and resistance to shortcut learning. GeoThinker and \name using configuration of Qwen3VL-8B with 32frame setting.
}
 \resizebox{1\linewidth}{!}{
\begin{tabular}{llcccc}
\toprule
\multirow{2}{*}{\textbf{Model}} & \multirow{2}{*}{\textbf{Benchmark}} & \multicolumn{4}{c}{\textbf{\# Frames}} \\
\cmidrule(lr){3-6}
 &  & \textbf{16} & \textbf{32} & \textbf{64} & \textbf{128} \\

\midrule
\multirow{2}{*}{Cambrian-S-7B}
& VSI & 58.6 & 63.6 & 66.4 & \sota{67.5} \\
& VSI-Debiased & 49.7 & 55.6 & 59.1 & \sota{59.9} \\
 
 
\midrule
\multirow{2}{*}{VG-LLM-8B*}
& VSI & 60.5 & 62.2 & \sota{63.7} & 63.1 \\
& VSI-Debiased & 51.6 & 52.4 & \sota{55.2} & 55.1 \\


\midrule
\multirow{2}{*}{GeoThinker}
& VSI & 69.2 & 72.6 & 73.4 & \sota{73.4} \\
& VSI-Debiased & 64.3 & 66.3 & 67.7 & \sota{68.1} \\

\midrule
\multirow{2}{*}{\name}
& VSI & 67.0 & 70.4 & 72.1 & \sota{72.4} \\
& VSI-Debiased & 61.7 & 65.8 & \sota{67.1} & 66.7 \\

\bottomrule
\end{tabular}}
  \label{tab:ablation-study-data}
\end{minipage}
\end{minipage}
\subsection{Embodied Intelligence}

To further demonstrate the efficiency advantage of \name in high-frequency, real-time control, we analyze the computational and latency profile on the LIBERO robotic manipulation benchmark under the StarVLA framework in Table~\ref{tab:appendix_libero_latency}.

\paragraph{Action-Chunking and Token Setup.}
Evaluations are conducted on a Qwen3-VL-4B backbone using Optimal Fine-Tuning following StarVLA-OFT. Unlike general VLM text generation, embodied VLA policies operate in a tight closed loop. 
In a representative setup, the model predicts a single action chunk of $8$ continuous action tokens with a compact input context of $241$ tokens, which comprises $128$ image tokens derived from $2$ camera views with $64$ tokens per view and $113$ language instruction tokens.

\paragraph{Latency Reduction in Short-Horizon Generation.}
In contrast to long-text generation scenarios where pre-processing latency can be amortized across hundreds of autoregressive decoding steps, robotic control policies generate very short output sequences with $8$ tokens. Consequently, the one-time feature extraction and prior synthesis overhead dominates overall end-to-end execution time. 

As shown in Table~\ref{tab:appendix_libero_latency}, maintaining an explicit external geometry encoder in GeoThinker introduces a massive computational burden: extra parameters surge to $1,705.5\,\text{M}$, and extra FLOPs jump by $+139.34\%$ ($3.40\,\text{T}$ on top of the $2.44\,\text{T}$ backbone). 
This heavy pre-processing pipeline causes execution latency to skyrocket from $82.0\,\text{ms}$ of raw QwenOFT to $254.8\,\text{ms}$, incurring an auxiliary prior latency overhead of $172.8\,\text{ms}$ that severely bottlenecks real-time control frequencies.


\paragraph{Lightweight Endogenous Prior Generation.}
By internalizing 3D perception via lightweight LFP heads, \name compresses the extra parameter footprint to just $277.0\,\text{M}$ and reduces extra FLOPs to a negligible $0.07\,\text{T}$, representing a \textbf{97.94\% reduction in additional compute} ($3.40\,\text{T} \rightarrow 0.07\,\text{T}$) and adding a mere $+2.86\%$ overhead over the raw backbone.

Crucially, \name slashes the additional prior integration latency from $172.8\,\text{ms}$ down to $22.2\,\text{ms}$, achieving an \textbf{87.15\% reduction in auxiliary latency}. This brings overall end-to-end execution time down from $254.8\,\text{ms}$ to $104.2\,\text{ms}$ with a \textbf{2.45$\times$ total speedup}, closing nearly $87\%$ of the latency gap back to the unaugmented backbone at $82.0\,\text{ms}$. These results highlight that for low-latency, short-horizon embodied control, \name provides an architecture that unlocks fine-grained 3D spatial intuition with virtually negligible computational overhead.


\section{General video mixture and test-frame scaling}



\subsection{Impact of General Video Data Mixture}
\label{supp:video_mixture_analysis}

To investigate how joint training on general video data affects specialized spatial reasoning, we evaluate \name against Cambrian-S and GeoThinker across VSI-Bench, VideoMME, and MVBench in Table~\ref{tab:ablation-study-structure}. 

\paragraph{Elevated Performance Baseline and Reduced Task Interference.}
As shown in Table~\ref{tab:ablation-study-structure}, pure 2D VLM architectures like Cambrian-S-7B suffer from severe task interference: incorporating 3M general video samples drops its VSI score by $4.1$ points ($69.2 \rightarrow 65.1$). In contrast, \name exhibits strong structural robustness. Even \textit{without} general video mixtures ($\boldsymbol{\mathcal{X}}$), \name achieves significantly higher baseline lower bounds on general video benchmarks (VideoMME: $55.9$, MVBench: $55.8$) compared to GeoThinker ($53.7$ and $42.8$). This indicates that endogenously learning 3D geometric and temporal dynamic prediction heads intrinsically enhances the backbone's underlying temporal representation capability.

\paragraph{Knowledge Absorption and Stage-1 Pretraining Bottlenecks.}
When incorporating a lightweight mixture of $430\text{k}$ general video samples, \name successfully absorbs new dynamic knowledge, yielding consistent gains across both general video tasks with +2.8 on VideoMME, +1.1 on MVBench and spatial reasoning with +0.4 on VSI. However, compared to GeoThinker's substantial performance leaps (+5.7 on VideoMME, +26.3 on MVBench), the relative gains for \name are more moderate. We hypothesize that this stems from the data volume in Stage-1 LFP pretraining: because the LFP heads are pretrained on a relatively compact geometry dataset, their endogenous feature generators incur minor prediction noise when generalizing to broader, out-of-distribution general video domains. Expanding Stage-1 LFP pretraining scale represents a promising avenue for further unlocking general video capabilities.




\subsection{Zero-Shot Frame Scaling and Extrapolation Limits}
\label{supp:frame_scaling_analysis}

To evaluate whether endogenously synthesized spatial-temporal priors generalize to longer temporal contexts, we evaluate models trained on $32$-frame windows by scaling inference frames up to 128 frames on VSI and VSI-Debiased benchmarks in Table~\ref{tab:ablation-study-data}.

\paragraph{Comparison with Passive Fusion Architectures.}
With our Spatial-Grounded Temporal (SGT) active perception mechanism, \name demonstrates clear advantages over conventional passive fusion baselines like VG-LLM-8B which plateaus at $63.7$ on VSI and degrades beyond 64 frames on VSI-Debiased. On standard VSI-Bench, \name scales smoothly from 16 to 128 frames ($67.0 \rightarrow 70.4 \rightarrow 72.1 \rightarrow 72.4$), proving that active feature routing effectively handles expanding sequence lengths without suffering from token clutter.

\paragraph{Insight on Endogenous Prediction Drift vs. Explicit 3D Anchors.}
An intriguing phenomenon emerges when evaluating on VSI-Debiased: while the explicit-encoder GeoThinker continues to gain accuracy up to 128 frames at $68.1$, \name peaks at 64 frames at $67.1$ and experiences a minor drop at 128 frames at $66.7$. This contrast reveals fundamental differences between explicit and endogenous prior generation:
\begin{enumerate}
    \item \textbf{Explicit World Anchors (GeoThinker):} The persistent runtime external VGGT encoder provides deterministic 3D spatial understanding across arbitrarily long multi-frame sequences, acting as an immutable physical world anchor.
    \item \textbf{Endogenous Prediction Drift (\name):} \name relies on self-predicted latent features synthesized by lightweight LFP heads. While these endogenous representations are accurate over short and medium contexts up to 64 frames, subtle feature prediction errors inevitably compound over extreme context lengths like 128 frames, leading to minor representation drift that slightly bounds long-sequence extrapolation on strictly debiased tasks.
\end{enumerate}
\name peaks at 64 frames and shows a modest decline at 128 frames on VSI-Debiased, suggesting sensitivity to longer-context extrapolation under the evaluated setting.

\section{Evaluation on VSTI-Bench}
\label{supp:vsti_bench}

\input{tables/11_vstibench}


To assess spatial reasoning in dynamic multi-view video scenes, we evaluate \name on VSTI-Bench~\citep{fan2025vlm(vlm3r)}. As reported in Table~\ref{tab:appendix_vstibench}, \name\textsubscript{Qwen3VL-8B} attains an average accuracy of $64.0\,\%$, exceeding several general open-source VLMs, including InternVL2-8B ($43.5\,\%$), as well as proprietary API models such as GPT-4o ($38.2\,\%$) and Gemini-1.5 Flash ($32.1\,\%$). It also outperforms the specialized 3D-augmented VLM-3R-7B by $5.2$ points ($64.0\,\%$ versus $58.8\,\%$), while avoiding runtime execution of an external geometry encoder. While Cambrian-P ($68.9\,\%$) and the external-encoder GeoThinker counterpart ($67.4\,\%$) obtain higher absolute scores, \name retains competitive performance with a substantially lighter inference architecture. Overall, these results indicate that endogenously generated geometry and future-semantic representations can support spatial reasoning on challenging multi-view video inputs.

\section{Evaluations on EASI Leaderboard}
\label{supp:easi_leaderboard}

To evaluate holistic spatial intelligence across standard and challenging benchmarks, we report results on the EASI leaderboards in Table~\ref{tab:easi_8_leaderboard} for EASI-8~\citep{cai2025holistic} and Table~\ref{tab:easi_all_leaderboard} for EASI-ALL~\citep{cai2025holistic}.

\input{tables/19_EASI_8_leaderboard}

\paragraph{Superior Advancement on EASI-8.}
As shown in Table~\ref{tab:easi_8_leaderboard}, \name\textsubscript{Qwen3VL-8B} achieves a jump on the EASI-8 leaderboard, advancing from the original GeoThinker's rank of 12th at AVG 59.1\% all the way to \textbf{2nd place globally} with an overall average score of \textbf{64.8\%}. Furthermore, \name surpasses leading proprietary foundation models, including Gemini 3 Pro at (63.8\%, 6th), GPT-5.2 at (59.5\%, 11th), and Gemini 2.5 Pro (58.0\%, 16th), as well as strong open-source spatial models like Qwen3.5-VL-35B at (61.6\%, 8th) and Cambrian-S-7B at (45.1\%, 36th). Notably, \name sets top-tier benchmarks in MindCube at $93.2\%$ and ViewSpatial at $82.6\%$. 
These results show that \name can achieve strong EASI-8 performance while eliminating runtime dependence on an external 3D geometry encoder.

\input{tables/10_EASI_ALL_leaderboard}

\paragraph{Performance on EASI-ALL.}
Table~\ref{tab:easi_all_leaderboard} extends the evaluation to the broader EASI-ALL suite, which covers diverse multimodal and temporal spatial reasoning settings. \name maintains competitive performance across these benchmarks, complementing the EASI-8 results reported in Table~\ref{tab:easi_8_leaderboard}. The results suggest that the proposed endogenous predictive representations can transfer beyond individual spatial reasoning tasks without requiring runtime execution of an external geometry encoder.


\section{Evaluation on ReVSI}
\label{supp:revsi_eval}

\subsection{Evaluation on ReVSI Benchmark}

\input{tables/14_appendix_revsi}

To rigorously evaluate spatial reasoning under dynamic visual conditions, we evaluate on ReVSI, a recently introduced benchmark that updates ground-truth answers as the evaluated frame count changes.

\paragraph{Performance Analysis and Comparison.}
As shown in Table~\ref{tab:appendix_revsi}, the explicit multi-encoder GeoThinker achieves the strongest performance among the evaluated open-source models, reaching an average score of $55.9$ by retaining access to geometry features from a runtime external encoder. In contrast, \name eliminates this runtime geometry branch while attaining an average score of $50.6$. It outperforms strong 2D spatial baselines such as Cambrian-S at $49.1$ and Cambrian-P without pose input at $50.1$, while remaining within $1.4$ points of Cambrian-P at $52.0$. These results demonstrate a competitive accuracy--efficiency trade-off, although the explicit geometry encoder retained by GeoThinker remains advantageous for maximizing ReVSI accuracy.


\paragraph{Distance-Centric Error Analysis.}
The sub-category breakdown shows that the largest differences between \name and GeoThinker occur in Absolute Distance and Relative Distance estimation. Specifically, \name obtains $65.2\,\%$ and $45.9\,\%$ on these two categories, compared with $74.7\,\%$ and $61.8\,\%$ for GeoThinker, corresponding to gaps of $9.5$ and $15.9$ points, respectively. In contrast, the differences on Object Count, Object Size, Room Size, and Relative Direction are substantially smaller. This concentration of errors on distance-centric questions suggests that precise metric estimation remains challenging when geometry-related representations are predicted internally rather than directly extracted by a dedicated geometry encoder. Improving metric accuracy and sensitivity therefore remains an important direction for endogenous spatial representations.

\subsection{Frame Scaling and Extrapolation on ReVSI}
\label{supp:revsi_scaling}

\input{tables/15_appendix_revis_frame}
We further examine zero-shot temporal extrapolation on ReVSI by scaling the inference frame count from 16 to 128 under dynamically adjusted ground-truth annotations, as reported in Table~\ref{tab:appendix_revis_frame}. \name improves from $48.6\,\%$ at 16 frames to $49.8\,\%$ at its 32-frame training budget and further reaches $50.4\,\%$ and $50.6\,\%$ at 64 and 128 frames, respectively. These results demonstrate positive zero-shot scaling beyond the training context, although the gains begin to saturate after 64 frames. GeoThinker improves from $52.6\,\%$ to $55.9\,\%$ over the same range and remains stronger at every frame budget, with the gap increasing from $4.0$ points at 16 frames to $5.3$ points at 128 frames. This trend suggests that endogenous representations benefit from additional temporal context but currently exploit very long sequences less effectively than persistent features produced by an explicit geometry encoder.



\section{Ablation study on OmniSpatial training data}

\input{tables/16_appendix_all_no_omnispatial}
\label{supp:omnispatial_ablation}

In the main scaled regime experiments, \name incorporates the OmniSpatial-Train dataset mixture, which drives substantial performance leaps on the ViewSpatial benchmark with $42.1\% \rightarrow 83.2\%$ on Qwen2.5VL-7B and $44.7\% \rightarrow 82.6\%$ on Qwen3VL-8B. 
To isolate the architectural contribution from the effect of additional OmniSpatial-Train data, we conduct a controlled comparison using the same training split as GeoThinker.

\paragraph{Data-Controlled Performance Parity.}
As reported in Table~\ref{tab:appendix_no_omnispatial}, when trained under strictly identical data distributions without OmniSpatial-Train, \name achieves comparable performance under matched data with the explicit dual-encoder baseline, GeoThinker. Across the six spatial intelligence benchmarks, \name obtains an average score of \textbf{60.56} on Qwen2.5VL-7B vs. GeoThinker's $60.43$ and \textbf{62.21} on Qwen3VL-8B vs. GeoThinker's $62.23$. 

\paragraph{Controlled Comparison with the External-Encoder Baseline.}
This rigorous controlled comparison delivers a empirical conclusion: the performance of \name is not only an artifact of favorable training data mixtures. When controlling for data variables, internalizing 3D perception via lightweight Latent Frame Prediction (LFP) heads 
achieves comparable performance under matched training data.
\name matches the spatial reasoning capacity of multi-encoder architectures while successfully shedding their heavy runtime execution bottlenecks.

\section{Ablation Study on Stage-2 Training}

\input{tables/17_appendix_stage3_ablation}
\label{supp:stage2_ablation}

To determine the optimal configuration for Stage-2 instruction tuning, we conduct systematic ablations on a Qwen2.5-VL-7B backbone in Table~\ref{tab:appendix_stage2_ablation}, isolating three key training variables: input frame budget (8 vs. 32 frames), inclusion of General Video data mixtures, and whether endogenous LFP heads are frozen or jointly fine-tuned during Stage 2.

\paragraph{Efficacy of Frame Budget and General Video Integration.}
Comparing row configurations in Table~\ref{tab:appendix_stage2_ablation} confirms that both frame scaling and general video data mixtures contribute positively to spatial reasoning. Expanding the training budget from 8 to 32 frames consistently elevates VSI accuracy with $63.6 \rightarrow 65.2$, while introducing general video data further enhances cross-benchmark performance with $65.55 \rightarrow 67.15$ with frozen LFP and $66.47 \rightarrow 66.84$ with unfrozen LFP, verifying that broad temporal diversity reinforces spatial perception.

\paragraph{Frozen vs. Fine-tuned LFP Heads.}
An interesting phenomenon emerges when evaluating LFP head status during Stage 2 in Scaled Regime:
\begin{enumerate}
    \item \textbf{Frozen LFP  (\xmark):} When trained with general video data, keeping LFP heads frozen yields a slightly higher VSI score at $67.6$ and a higher overall Cross-Benchmark Average at $67.15$. However, this setup suffers a notable drop on MMSI-Bench at $28.1$, indicating a trade-off between specialized spatial performance and multimodal video reasoning.
    \item \textbf{Jointly Fine-tuned LFP (\cmark):} Unfreezing LFP heads allows endogenous feature synthesis to co-adapt dynamically with the MLLM backbone on instruction data. This joint optimization drastically improves MMSI performance to \textbf{31.2}, narrowing the gap to GeoThinker to just $-0.5$, while preserving strong performance on VSI on $67.3$ ($-1.2$ vs. GeoThinker) and maintaining an overall Cross-Benchmark Average of \textbf{66.84}.
\end{enumerate}

\paragraph{Configuration Selection.}
Although freezing LFP heads yields marginally higher scores on isolated spatial benchmarks, we select joint LFP fine-tuning with general video mixtures as our final default strategy. This configuration achieves the most balanced, holistic spatial-temporal intelligence profile without sacrificing multi-modal generalization. 
Crucially, all 32-frame \name variants based on Qwen2.5-VL-7B sustain an ultra-low inference latency of \textbf{0.48\,s}, demonstrating a $3.0\times$ acceleration over GeoThinker at $1.45\,\text{s}$ regardless of Stage-2 parameter choices. With the Qwen3-VL-8B backbone, \name further reduces latency from $1.45\,\text{s}$ to \textbf{0.40\,s}, yielding a $3.6\times$ speedup.

\section{Stage 1 Pre-training Loss Function}






\subsection{Basic Operations}
We define several elementary operations used throughout our framework. 
Given a vector $x \in \mathbb{R}^D$, its $L_2$-normalized variant is denoted by $\hat{x} = x / \|x\|_2$, and its element-wise logarithmic amplitude transformation is defined as $\log(1 + |x|)$. 
For a feature matrix $X = (x_{ij}) \in \mathbb{R}^{N \times D}$, the frame-level mean $\mu_x^{(i)}$ and standard deviation $\sigma_x^{(i)}$ across feature dimensions are computed as $\mu_x^{(i)} = \frac{1}{D}\sum_{j=1}^{D} x_{ij}$ and $\sigma_x^{(i)} = \sqrt{\frac{1}{D}\sum_{j=1}^{D}(x_{ij} - \mu_x^{(i)})^2}$, respectively. 
Furthermore, the cosine similarity between two vectors $x$ and $y$ is computed via $\cos(x, y) = (x \cdot y) / (\|x\|_2 \|y\|_2)$.

To handle regression targets robustly, we adopt the SmoothL1 Loss, defined as:
\begin{equation}
\mathrm{SmoothL1}(x, y) =
\begin{cases}
\frac{1}{2}(x - y)^2, & \text{if } |x - y| \le 1, \\[3pt]
|x - y| - \frac{1}{2}, & \text{otherwise.}
\end{cases}
\end{equation}







\subsection{Adjacent Difference Loss}
\label{sec:appendix_local_relation_loss}
Standard element-wise metrics like $L_1$ loss or cosine similarity~\citep{yang2025cambrian,assran2025v(V-JEPA2),maes2026leworldmodel(LeWorldModel)} primarily operate on individual tokens in isolation. Consequently, they often fail to capture the underlying temporal or structural trends across consecutive frames, leading to inconsistent predictions over long sequences. To enforce local structural smoothness and trend consistency, we introduce the Local Relation Loss $\mathcal{L}_{\mathrm{rel}}$, which explicitly penalizes local discrepancies in both direction and magnitude between predicted feature sequences $p = (p_1, \dots, p_S)$ and ground-truth features $g = (g_1, \dots, g_S)$.

Specifically, we first compute the first-order temporal differences for both predictions and ground truths to represent local dynamics:
\begin{equation}
    \Delta p_i = p_{i+1} - p_i, \qquad
    \Delta g_i = g_{i+1} - g_i, \qquad i = 1, \dots, S-1.
\end{equation}

To constrain the directional consistency of local transitions, the directional relation loss $\mathcal{L}_{\mathrm{adj}}^{\mathrm{dir}}$ is formulated as the mean squared error over the $L_2$-normalized difference vectors:
\begin{equation}
    \mathcal{L}_{\mathrm{adj}}^{\mathrm{dir}} = \frac{1}{S-1}\sum_{i=1}^{S-1} \big\| \widehat{\Delta p}_i - \widehat{\Delta g}_i \big\|_2^2,
\end{equation}
where $\widehat{\Delta p}_i$ and $\widehat{\Delta g}_i$ denote the $L_2$-normalized variants of $\Delta p_i$ and $\Delta g_i$, respectively.

Simultaneously, to penalize local amplitude variations without being overwhelmed by global scale fluctuations, we define a magnitude-aware scale factor $s_{\Delta} = \max\!\left(\frac{1}{S-1}\sum_{i=1}^{S-1} |\Delta g_i|,\ \varepsilon\right)$, where $\varepsilon$ is a small constant for numerical stability. The magnitude loss $\mathcal{L}_{\mathrm{adj}}^{\mathrm{mag}}$ applies a logarithmic compression combined with SmoothL1 loss:
\begin{equation}
    \mathcal{L}_{\mathrm{adj}}^{\mathrm{mag}} = \frac{1}{S-1}\sum_{i=1}^{S-1} \mathrm{SmoothL1}\!\left(
    \log\!\Big(1 + \frac{|\Delta p_i|}{s_{\Delta}}\Big),\ 
    \log\!\Big(1 + \frac{|\Delta g_i|}{s_{\Delta}}\Big)
    \right).
\end{equation}

The overall Local Relation Loss is a weighted combination of the directional and magnitude objectives:
\begin{equation}
    \mathcal{L}_{\mathrm{adj}}(p, g) = \mathcal{L}_{\mathrm{adj}}^{\mathrm{dir}} + \lambda_{\mathrm{mag}} \mathcal{L}_{\mathrm{adj}}^{\mathrm{mag}},
\end{equation}
where we empirically set $\lambda_{\mathrm{mag}} = 0.5$ in our experiments.

\subsection{Temporal Branch}
The temporal branch aims to model transition dynamics by training the Vision-Language Model (VLM) to predict the image hidden states of the next frame based on the current frame's representations. 

\textbf{Motivation and Supervision Scheme.}
Instead of employing redundant pixel-level supervision, we directly supervise predictions in the feature space. 
While self-supervised representation models (e.g., V-JEPA~\citep{assran2025v(V-JEPA2)}) or auxiliary temporal regularizers (e.g., Cambrian-S~\citep{yang2025cambrian}) rely effectively on simple $L1$ loss or cosine similarity, such a single-metric objective is insufficient for our fully supervised feature prediction target in our setting. 
Because image hidden states typically possess extremely small numerical scales, lightweight or naive loss functions tend to induce trivial shortcuts, leading the optimization toward premature false convergence. 
To guarantee precise feature alignment across scale, direction, statistical distribution, and dynamic trajectory, we propose a multi-faceted objective combining five complementary constraints.

\textbf{Branch Activation Gating.}
Since next-frame ground-truth features are non-existent in single-frame inputs, the temporal branch is conditionally activated using an indicator-based weighting mechanism:
\begin{equation}
    \lambda_{\mathrm{sem}}^{\mathrm{eff}} = \lambda_{\mathrm{sem}} \cdot \mathbb{1}\!\left[\, T \ge T_{\mathrm{threshold}} \ \wedge\ T > 1 \,\right],
\end{equation}
where $T$ denotes the input sequence length, and $T_{\mathrm{threshold}}$ represents the sequence length threshold required to initiate temporal training.

\paragraph{Direction Loss.}
To align the spatial orientation of predicted hidden states $p_i$ with ground-truth next-frame features $g_i$, we minimize the $L_2$ distance between their $L_2$-normalized vectors across $M$ samples:
\begin{equation}
    \mathcal{L}_{\mathrm{sem}}^{\mathrm{dir}} = \frac{1}{M}\sum_{i=1}^{M} \big\| \hat{p}_i - \hat{g}_i \big\|_2^2.
\end{equation}

\paragraph{Relative Amplitude Loss.}
To supervise magnitude scale without being dominated by feature outliers, we normalize absolute values by an adaptive sequence baseline $s = \max\!\left(\frac{1}{M}\sum_{i=1}^{M} |g_i|,\ \varepsilon\right)$ and apply logarithmic dynamic compression:
\begin{equation}
    \mathcal{L}_{\mathrm{sem}}^{\mathrm{rel\text{-}mag}} = \frac{1}{M}\sum_{i=1}^{M} \mathrm{SmoothL1}\!\left(
    \log\!\Big(1 + \frac{|p_i|}{s}\Big),\ 
    \log\!\Big(1 + \frac{|g_i|}{s}\Big)
    \right).
\end{equation}

\paragraph{Statistic Loss.}
To encourage matching in global feature distributions, we penalize the discrepancies between the channel-wise means ($\mu_p^{(i)}, \mu_g^{(i)}$) and standard deviations ($\sigma_p^{(i)}, \sigma_g^{(i)}$):
\begin{equation}
    \mathcal{L}_{\mathrm{sem}}^{\mathrm{stats}} = \frac{1}{M}\sum_{i=1}^{M} \mathrm{SmoothL1}\big(\mu_p^{(i)},\, \mu_g^{(i)}\big) + \frac{1}{M}\sum_{i=1}^{M} \mathrm{SmoothL1}\big(\sigma_p^{(i)},\, \sigma_g^{(i)}\big).
\end{equation}

\paragraph{Temporal Rank Loss.}
To ensure the predicted feature $p_i$ moves strictly toward the target next-frame feature $g_i$ rather than remaining static near the current-frame feature $\tilde{g}_i$, we enforce a pairwise margin constraint:
\begin{equation}
    \mathcal{L}_{\mathrm{sem}}^{\mathrm{rank}} = \frac{1}{M}\sum_{i=1}^{M} \max\!\left(0,\ \big(1 - \cos(p_i, g_i)\big) - \big(1 - \cos(p_i, \tilde{g}_i)\big) + m\right),
\end{equation}
where $m = 0.05$ represents the distance margin, and $\tilde{g}_i$ denotes the ground-truth feature of the current frame.

\paragraph{Adjacent Difference Loss.}
Finally, to preserve structural continuity across consecutive steps, we incorporate the local relation loss defined in Sec.~\ref{sec:appendix_local_relation_loss}:
\begin{equation}
    \mathcal{L}_{\mathrm{sem}}^{\mathrm{adj}} = \mathcal{L}_{\mathrm{adj}}\big(p,\ g\big).
\end{equation}

\textbf{Overall Objective.} The overall loss for the temporal branch is formulated as a weighted sum:

\begin{equation}
    \mathcal{L}_{\mathrm{sem}} = \mathcal{L}_{\mathrm{sem}}^{\mathrm{dir}} + 0.5\, \mathcal{L}_{\mathrm{sem}}^{\mathrm{rel\text{-}mag}} + 0.15\, \mathcal{L}_{\mathrm{sem}}^{\mathrm{stats}} + 0.15\, \mathcal{L}_{\mathrm{sem}}^{\mathrm{rank}} + 0.2\, \mathcal{L}_{\mathrm{sem}}^{\mathrm{adj}}.
\end{equation}

\subsection{Geometry Branch}
To endow the model with global 3D geometric awareness from a single current frame, the geometry branch learns to map the VLM's current image hidden states $\mathrm{geo\_src}$ into global VGGT feature representations. Unlike the temporal branch, this prediction operates purely in the spatial domain without temporal dynamics. 
Although each target is spatially aligned with frame $(i)$, it is produced through joint multi-view encoding and should therefore be interpreted as a current-view-anchored representation of scene-level geometry, rather than a feature derived from frame $(i)$ alone.

\textbf{Feature Targets and Fallback Guidance.}
The predicted geometry representations $p = \phi_{\mathrm{geo}}(\mathrm{geo\_src})$ are supervised by target features $g$, extracted by a pre-trained VGGT encoder fed with multi-view context. To accommodate datasets where ground-truth geometry tokens ($\mathrm{geo\_tokens}$) are partially missing, we employ a stop-gradient fallback mechanism to avoid gradient contamination:
\begin{equation}
    p = \phi_{\mathrm{geo}}\big(\mathrm{geo\_src}\big), \qquad
    g = \begin{cases}
    \mathrm{geo\_tokens}, & \text{if GT geotokens are available,} \\[2pt]
    \mathrm{sg}\big(p\big), & \text{otherwise,}
    \end{cases}
\end{equation}
where $\mathrm{sg}(\cdot)$ denotes the stop-gradient operator. When ground truth is unavailable, $g = \mathrm{sg}(p)$ ensures the loss evaluates to zero, naturally bypassing gradient updates for non-geometric samples.

\textbf{Cross-Modal Feature Alignment.}
Due to structural disparities between VLM representations and VGGT feature spaces, directly enforcing element-wise alignment easily leads to optimization instability. The geometry loss $\mathcal{L}_{\mathrm{geo}}$ incorporates directional alignment $\mathcal{L}_{\mathrm{geo}}^{\mathrm{dir}}$, relative magnitude regularization $\mathcal{L}_{\mathrm{geo}}^{\mathrm{rel\text{-}mag}}$, and adjacent difference loss $\mathcal{L}_{\mathrm{geo}}^{\mathrm{adj}}$ formulated identically to Sec.~\ref{sec:appendix_local_relation_loss}. 

To specifically address the drastic scale mismatch between VLM hidden states and VGGT features, we adapt the channel-wise statistic loss with explicit scale normalization:
\begin{equation}
    s_s = \max\big(\mathrm{sg}(s),\ \varepsilon\big), \qquad \text{where } s = \frac{1}{M}\sum_{i=1}^{M} |g_i|,
\end{equation}
\begin{equation}
    \mathcal{L}_{\mathrm{geo}}^{\mathrm{stats}} = \frac{1}{M}\sum_{i=1}^{M} \mathrm{SmoothL1}\!\Big(\frac{\mu_p^{(i)}}{s_s},\, \frac{\mu_g^{(i)}}{s_s}\Big) + \frac{1}{M}\sum_{i=1}^{M} \mathrm{SmoothL1}\!\Big(
    \log\!\Big(1 + \frac{\sigma_p^{(i)}}{s_s}\Big),\ 
    \log\!\Big(1 + \frac{\sigma_g^{(i)}}{s_s}\Big)
    \Big).
\end{equation}
Normalizing channel means $\mu^{(i)}$ by $s_s$ prevents large-magnitude dimensions from dominating the gradient, while log-compressing normalized standard deviations $\sigma^{(i)}$ stabilizes variance matching across divergent feature spaces.

\textbf{Overall Objective.} The combined loss for the geometry branch is formulated as:
\begin{equation}
    \mathcal{L}_{\mathrm{geo}} = \mathcal{L}_{\mathrm{geo}}^{\mathrm{dir}} + 0.5\, \mathcal{L}_{\mathrm{geo}}^{\mathrm{rel\text{-}mag}} + 0.15\, \mathcal{L}_{\mathrm{geo}}^{\mathrm{stats}} + 0.35\, \mathcal{L}_{\mathrm{geo}}^{\mathrm{adj}}.
\end{equation}

\section{Visualization}

\begin{figure}[t]
  \centering

  \begin{subfigure}[t]{\textwidth}
    \centering
    \includegraphics[width=\textwidth]{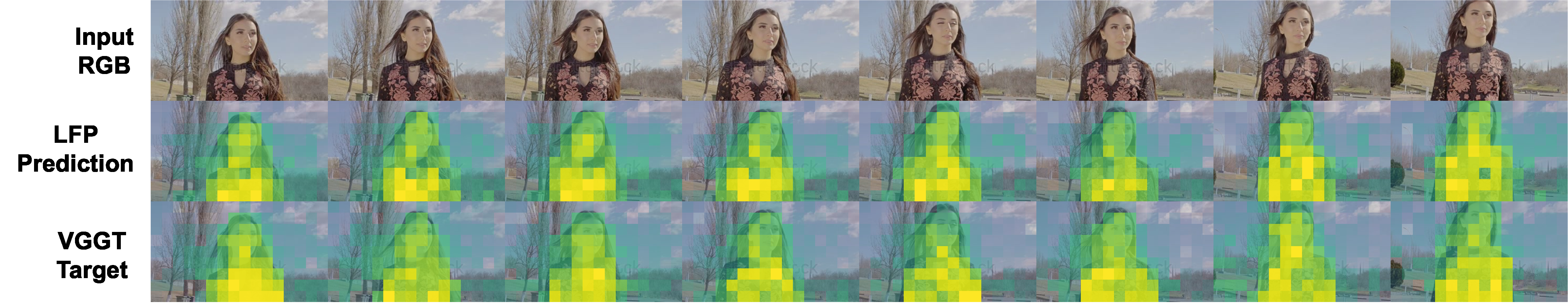} 
  \end{subfigure}
  \begin{subfigure}[t]{\textwidth}
    \centering
    \includegraphics[width=\textwidth]{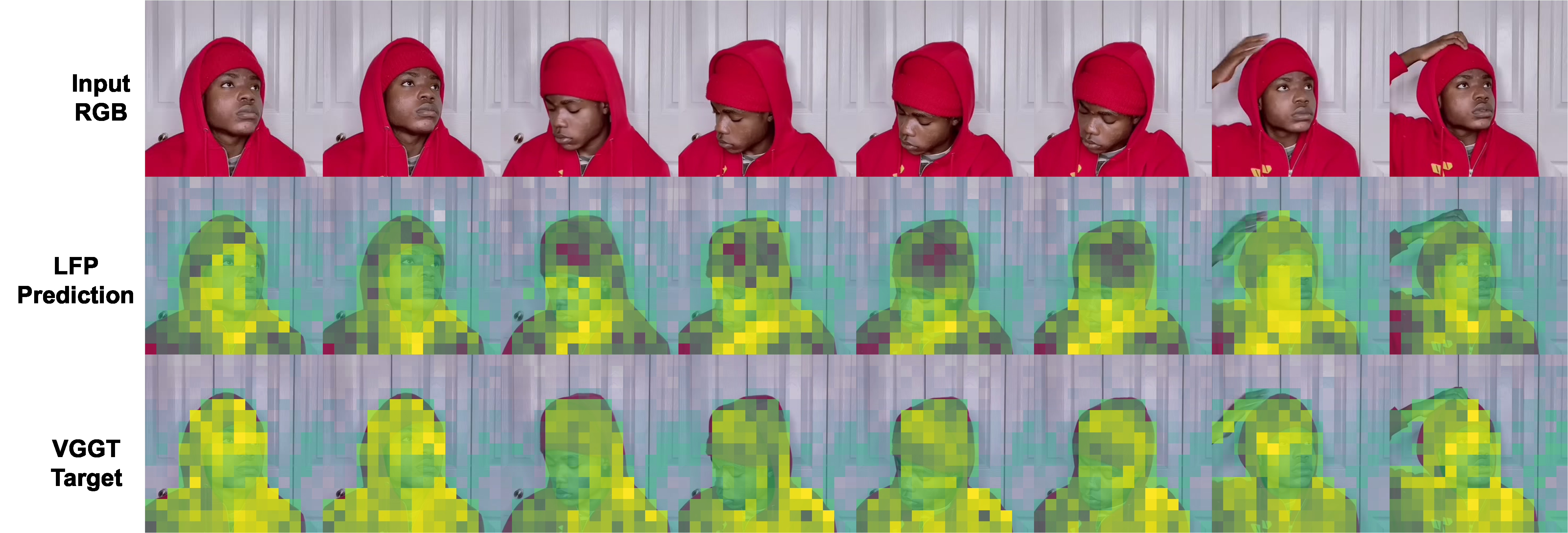} 
  \end{subfigure}
  \caption{
  \textbf{Qualitative structural correspondence in standard scenes.} For each sequence, we show input RGB frames (top), LFP-predicted geometry representations (middle), and the corresponding VGGT feature targets (bottom). In these selected scenes with clear foreground subjects and structured motion, the endogenous predictions preserve coherent object-centered activation patterns and broad spatial correspondence with the VGGT targets across consecutive frames. This qualitative agreement illustrates that \name\ can recover useful geometry-related structure from intermediate MLLM hidden states.
  }
  \label{fig:appendix_visualization_2}
\end{figure}

\begin{figure}[t]
  \centering

  \begin{subfigure}[t]{\textwidth}
    \centering
    \includegraphics[width=\textwidth]{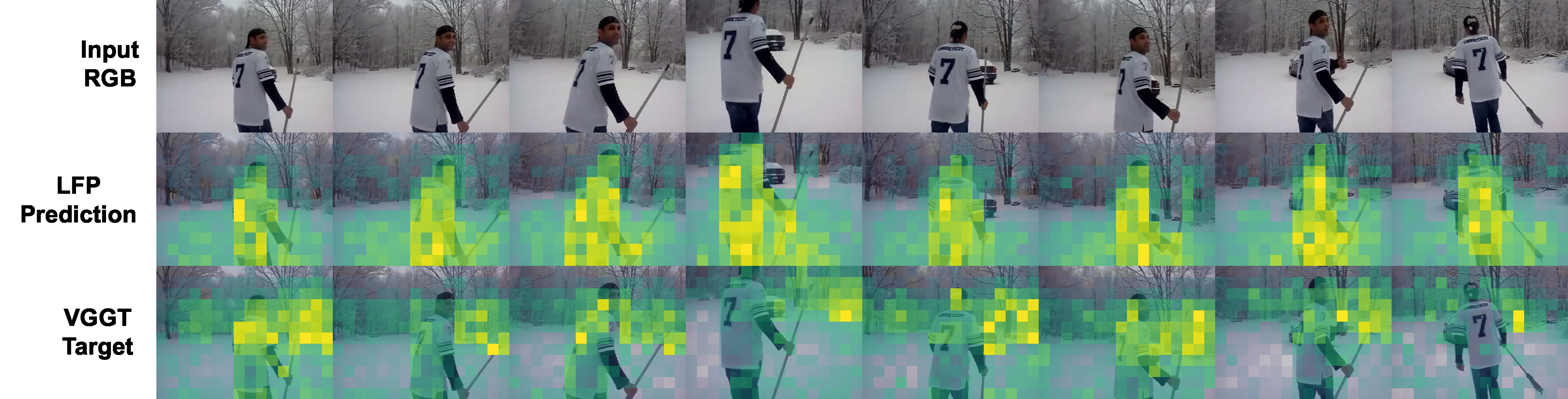} 
  \end{subfigure}
  \begin{subfigure}[t]{\textwidth}
    \centering
    \includegraphics[width=\textwidth]{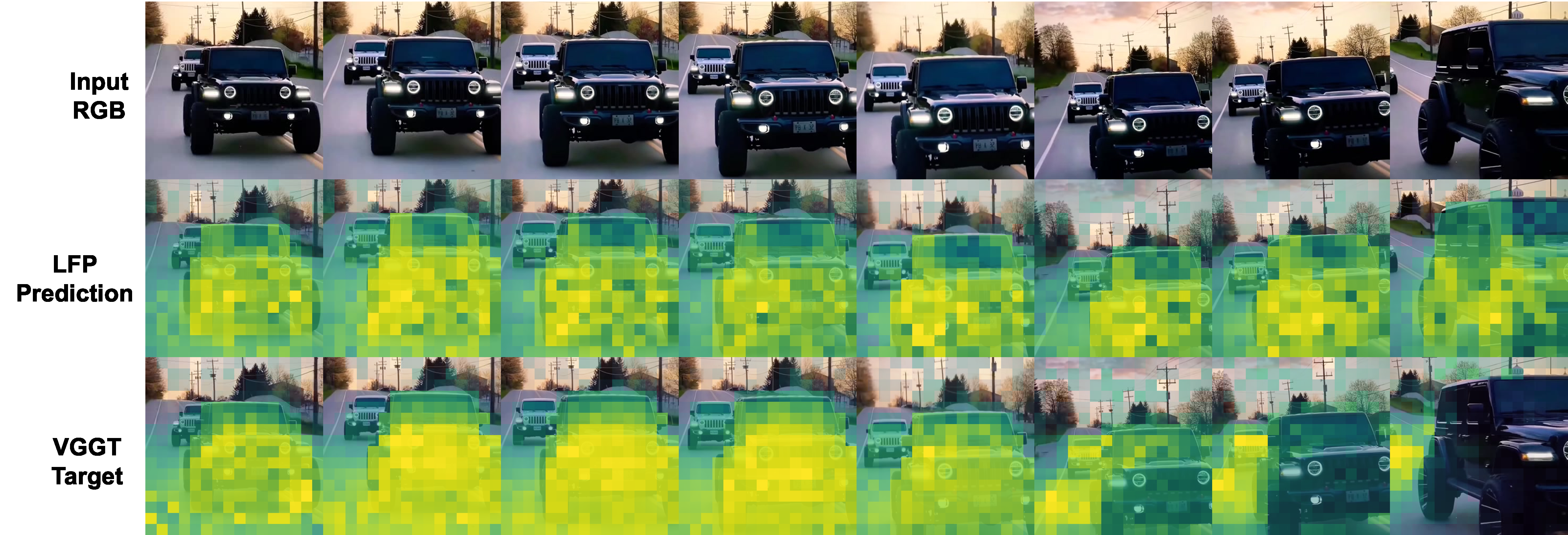} 
  \end{subfigure}
  \caption{
  \textbf{Qualitative semantic selectivity under visually ambiguous conditions.} For each sequence, we show input RGB frames (top), LFP-predicted geometry representations (middle), and the corresponding VGGT feature targets (bottom). In these selected snow and driving examples, the VGGT targets exhibit comparatively diffuse or background-dominated activations, whereas the endogenous predictions remain concentrated around the primary human or vehicle. This case-level contrast suggests that prediction from MLLM hidden states can preserve semantically relevant spatial focus when low-level visual cues are ambiguous.
  }
  \label{fig:appendix_visualization_3}
\end{figure}

\begin{figure}[t]
  \centering

  \begin{subfigure}[t]{\textwidth}
    \centering
    \includegraphics[width=\textwidth]{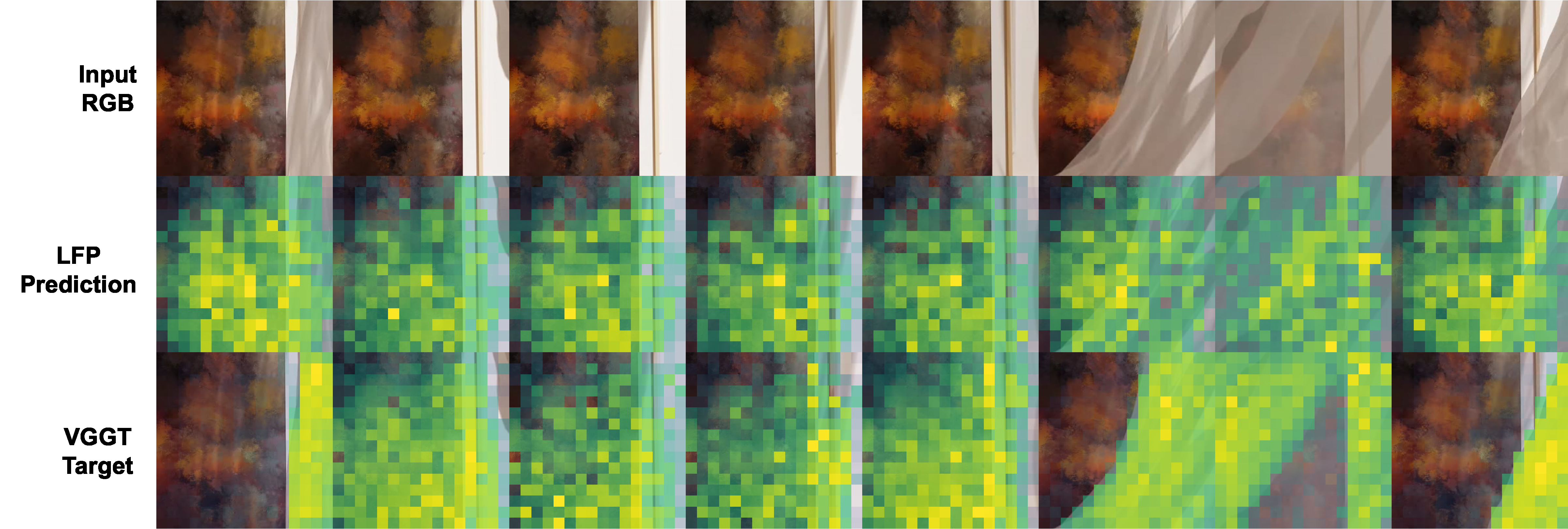} 
  \end{subfigure}
  \begin{subfigure}[t]{\textwidth}
    \centering
    \includegraphics[width=\textwidth]{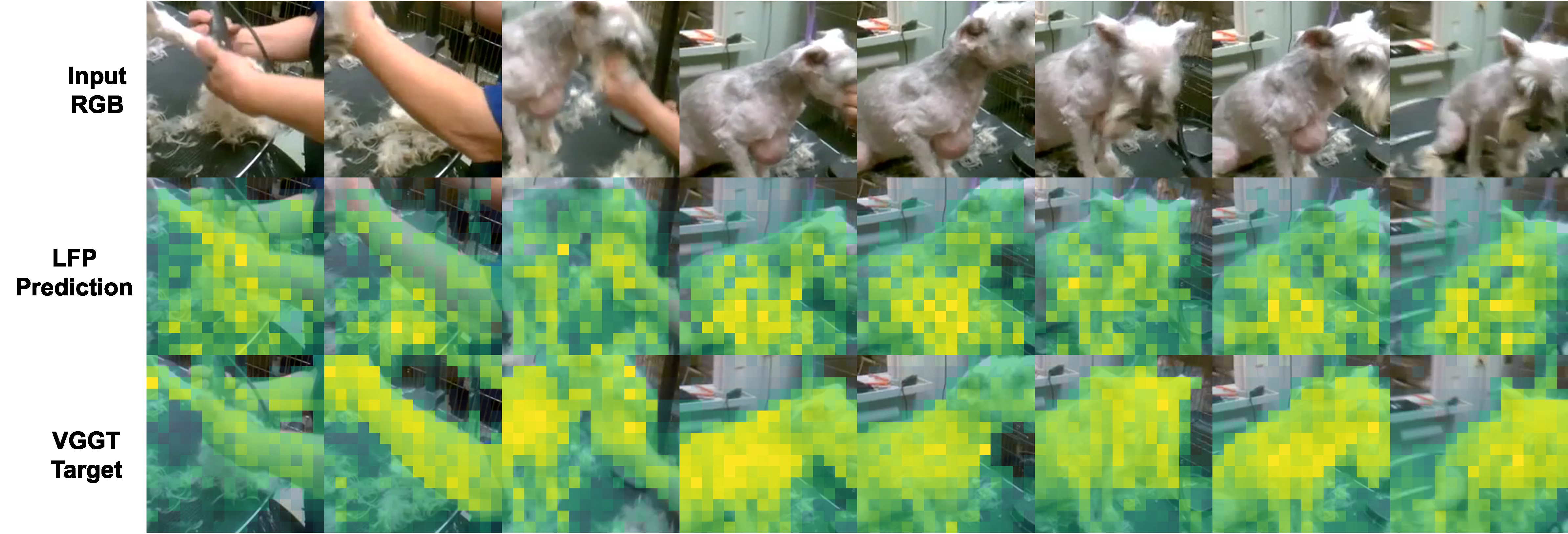} 
  \end{subfigure}
  \caption{
  \textbf{Qualitative limitations of endogenous geometry prediction.} For each sequence, we show the input RGB frames (top), LFP-predicted geometry representations (middle), and the corresponding VGGT feature targets (bottom). \textbf{Top:} In the presence of rapidly changing non-rigid structures, the predicted representations exhibit weaker correspondence with the VGGT targets. \textbf{Bottom:} The predicted maps display coarser spatial variation than the corresponding VGGT feature targets. These examples illustrate conditions in which endogenous geometry prediction does not fully recover the geometry-related structure captured by the external expert.
  }
  \label{fig:appendix_visualization_1}
\end{figure}

To qualitatively assess the spatial structure captured by our endogenously predicted geometry features, we visualize the outputs of the Latent Frame Prediction (LFP) heads alongside the corresponding VGGT target features across diverse scenarios. For each feature tensor, we average its activations along the channel dimension, min-max normalize the resulting 2D map to \([0,1]\), and render it using the same colormap. Accordingly, these visualizations compare spatial activation patterns rather than absolute feature magnitudes.

\paragraph{Structural Correspondence in Standard Scenarios.}
Figure~\ref{fig:appendix_visualization_2} qualitatively compares the predicted geometry representations with their corresponding VGGT feature targets in canonical scenes containing clear foreground subjects and structured motion. Across consecutive frames, the predicted maps preserve broad spatial activation patterns and object-level structures visible in the VGGT targets. These visualizations suggest that intermediate MLLM hidden states contain information that can be transformed into useful geometry-related representations, without requiring an exact reproduction of individual VGGT feature activations.

\paragraph{Qualitative Behavior under Ambiguous Backdrops.}
Figure~\ref{fig:appendix_visualization_3} further compares predicted geometry representations and VGGT feature targets in scenes with visually ambiguous backgrounds. In several low-contrast snow sequences, the predicted maps remain concentrated around the human subject, whereas the corresponding VGGT targets exhibit more diffuse activations. This qualitative difference is consistent with high-level semantic intuition information in the MLLM hidden states contributing to the predicted representations. A broader evaluation would be required to establish whether this behavior systematically improves robustness under ambiguous visual conditions.

\paragraph{Limitations and Failure Modes.} 
Despite these advantages, we observe two main limitations of endogenous feature prediction in Figure~\ref{fig:appendix_visualization_1}:
\begin{itemize}
    \item \textit{Transient Non-Rigid Dynamics}: As shown in the top row of Figure~\ref{fig:appendix_visualization_1}, when fine-grained, non-rigid dynamic objects appear (e.g., a translucent curtain fluttering in the wind), VGGT precisely captures the moving geometric boundary, whereas \name\ fails to register these transient structural shifts. 
    This failure may reflect the limited sensitivity of the predicted representations to fine-grained, rapidly changing visual structures.
    \item \textit{Representation Smoothness}: Compared to the dense, continuous outputs generated by dedicated 3D encoders, our self-predicted features are visibly coarser and display higher spatial jitter (Figure~\ref{fig:appendix_visualization_1}, bottom row). This graininess arises from the token-level spatial discretization inherent in Transformer architectures. Future work will explore smoothed decoder losses or continuous implicit fields to further improve feature regularity.
\end{itemize}

\clearpage
\section{Discussion and Scope}

We use \emph{predictive world representations} in a restricted operational sense: internally generated latent representations encoding current multi-view geometry and short-horizon future visual semantics. \name does not aim to construct an action-conditioned or long-horizon generative world model. Instead, it investigates whether explicit world representations previously supplied by an external geometry encoder can be predicted from causal MLLM states and reused for spatial reasoning.

We distinguish \name from both conventional feature distillation and latent-token distillation at the level of the inference interface rather than representational capacity. Conventional feature distillation typically absorbs teacher supervision into backbone parameters without exposing separately generated, scene-conditioned geometry priors at inference. Latent-token methods such as 3DThinker instead supervise compact geometry-related token states within an autoregressive reasoning trajectory, allowing geometric information to influence subsequent decoding implicitly. In contrast, \name predicts dense, frame-aligned geometry and future-semantic priors at selected intermediate layers and repeatedly integrates them into the backbone through an explicit fusion interface. We therefore do not claim that distilled models cannot encode or dynamically utilize geometric information. Our distinction concerns where and how geometry-related evidence is generated, exposed, and routed during inference. Accordingly, our empirical comparison characterizes the accuracy-efficiency trade-off under the reported settings rather than establishing an inherent limitation of distillation.

The reported acceleration arises from two complementary sources. On Qwen3-VL-8B, replacing runtime external geometry extraction with endogenous prediction reduces latency from 1.66 s to 0.91 s without further token compression, corresponding to a 1.82× architectural speedup. Combining endogenous prediction with spatial token compression reduces complete-system latency from 1.45 s to 0.40 s, yielding a 3.63× end-to-end speedup. These latency values are specific to the reported hardware and input configuration, while the corresponding FLOP and MAC reductions provide hardware-independent evidence of lower computational cost.

\name trades offline teacher supervision and two-stage training for lower repeated inference cost. It is therefore most suitable for streaming, interactive, and embodied settings where geometry features cannot be precomputed. Explicit geometry encoders may remain preferable when maximum metric accuracy is required, as reflected by the remaining gap on distance-centric ReVSI categories. Our goal is not to dominate external-encoder systems under every metric, but to provide a favorable accuracy-efficiency trade-off without runtime dependence on a separate geometry model.


%% file: tables/05_recogdrive.tex
\begin{table}[h]
    \centering
    \small
    \vspace{1mm}
    \caption{\textbf{Performance comparison on NAVSIM \textit{navtest}.}
    Compared with ReCogDrive, \name improves key driving metrics. Compared with GeoThinker, it reduces the spatial-prior integration overhead by 52.91\%.
    Evaluation includes \underline{N}ot-at-fault \underline{C}ollisions (NC), \underline{D}rivable \underline{A}rea \underline{C}ompliance (DAC), \underline{T}ime-\underline{T}o-\underline{C}ollision within bound (TTC), \underline{Comf}ort (Comf.), \underline{E}go and \underline{P}rogress (EP).
    }
    \setlength{\tabcolsep}{2pt}
    \begin{tabular}{@{}l|cc|cc|ccc|cc@{}}
        \toprule
        \multirow{2}{*}{\textbf{Method}} &
        \multirow{2}{*}{\textbf{FLOPs(T)}} & 
        \textbf{Amortized Latency} & 
        \multirow{2}{*}{\textbf{NC$\uparrow$}} & \multirow{2}{*}{\textbf{DAC$\uparrow$}} & 
        \multirow{2}{*}{\textbf{TTC$\uparrow$}} & 
        \multirow{2}{*}{\textbf{Comf.$\uparrow$}} & 
        \multirow{2}{*}{\textbf{EP$\uparrow$}} & 
        \multirow{2}{*}{\textbf{PDMS$\uparrow$}} \\

         &  &  (ms/output token) &  &  &  &  &  &  \\
        \midrule
        Constant Velocity & -- & -- &68.0 & 57.8 & 50.0 & 100 & 19.4 & 20.6 \\
        Ego Status MLP & -- & -- & 93.0 & 77.3 & 83.6 & 100 & 62.8 &  65.6 \\

        \midrule
        ReCogDrive w/ InternVL & 73.45 & 32.15 &  95.0 & 93.7 &	87.8 &	100 &	78.7 &	81.9 \\
        GeoThinker  & 122.69 & 42.62 &  98.0 &	93.2 &	94.1 &	100 &	78.4 &	\sota{84.4} \\
        \rowcolor{navyblue!10} 
        \name\textit{(Ours)} & 88.22 & 37.08\textcolor{blue}{(-52.91\%)} & 96.5 & 94.0 & 90.4 & 100 & 78.7 & 83.3 \\

        \bottomrule
    \end{tabular}
    \label{tab:comparison_recogdrive}
\end{table}

%% file: tables/20_libero.tex
\begin{table}[h!]
  \centering
  \caption{\textbf{Comparison of different VLA models on LIBERO.} 
  We train one policy for all 4 suites. All scores are averaged over 500 trials for each task suite (10 tasks × 50 episodes).
  Compared with GeoThinker, it reduces the spatial-prior integration overhead by 87.15\%.
  }
  \label{tab:apendix_libero_results}
  \setlength{\tabcolsep}{2pt}
  \begin{tabular}{l |c c|c c c c c c}
    \toprule
    \multirow{2}{*}{\textbf{Model}} & 
    \multirow{2}{*}{\textbf{FLOPs(T)}} & 
    \multirow{2}{*}{\textbf{Latency(ms)}} &
    \multirow{2}{*}{\textbf{Steps}} & 
    \multirow{2}{*}{\textbf{Spatial}} & 
    \multirow{2}{*}{\textbf{Object}} & 
    \multirow{2}{*}{\textbf{Goal}} & 
    \multirow{2}{*}{\textbf{Long}} & 
    \multirow{2}{*}{\textbf{Avg}} \\
    
     &  &  & &  &  &  &  &  \\
    \midrule
    \rowcolor{black!11}
    \multicolumn{9}{l}{\textbf{Baseline}} \\
    $\pi_0$+FAST & - & - & - & 96.4 & 96.8 & 88.6 & 60.2 & 85.5 \\
    $\pi_0$ & - & - & - & 96.8 & 98.8 & 95.8 & 85.2 & 94.1 \\
    OpenVLA-OFT-7B & - & - & 175K & 97.6 & 98.4 & 97.9 & 94.5 & \sota{97.1} \\
    GR00T-N1.5 & - & - & 20K  & 92.0 & 92.0 & 86.0 & 76.0 & 86.5 \\

    \midrule
    \rowcolor{black!11}
    \multicolumn{9}{l}{\textbf{Qwen3-VL-4B}} \\

    StarVLA-OFT & 2.44 & 82.0 & 30K & 97.8 & 98.6 & 96.2 & 93.8 & 96.6 \\

    GeoThinker & 5.84 & 254.8 & 30K & 98.4 & 99.0 & 98.4 & 91.2 & 96.7 \\
    \rowcolor{navyblue!10} 
    \name\textit{(Ours)} & 2.51 & 104.2\textcolor{blue}{(-87.15\%)}  & 30K & 97.6 & 99.6 & 98.2 & 93.0 & \sota{97.1} \\



    \bottomrule
  \end{tabular}
\end{table}

%% file: tables/09_appendix_flops.tex
\begin{table}[t]
\vspace{-1mm}
\centering
\small
\setlength{\tabcolsep}{0.5mm}
\caption{
\textbf{Quantitative efficiency and spatial reasoning performance comparison across different model backbones.} We report computational cost in FLOPs (T) and MACs (T), inference latency (s), and accuracy on VSI-Bench. * denotes leveraging estimated lower bound of latency. \name\textsubscript{32frame} achieves substantial reductions in FLOPs, MACs, and inference latency compared to external 3D encoder baselines (e.g., GeoThinker), while maintaining highly competitive performance.
}
\resizebox{0.8\linewidth}{!}{
\begin{tabular}{lccc|c}
\toprule
\textbf{Model} & \textbf{FLOPs(T)$\downarrow$} & \textbf{MACs(T)$\downarrow$} & \textbf{Latency(s)$\downarrow$} & \textbf{VSI$\uparrow$}\\

\midrule
\rowcolor{black!9}
\multicolumn{5}{l}{\textbf{Qwen2.5VL-3B}} \\
Qwen-2.5VL-3B & 95.29 & 47.64 & 0.60 & 28.6\\
3DThinker\textsubscript{Qwen-2.5VL-3B}* & 95.29 & 47.64 & 0.60 & 59.1\\
VG-LLM-4B & 152.92 & 76.43 & 1.43 & 46.7\\
GeoThinker\textsubscript{8frame} & 157.81 & 78.87 & 1.48 & 48.9\\

\midrule
\rowcolor{black!9}
\multicolumn{5}{l}{\textbf{Qwen2.5VL-7B}} \\

Qwen-2.5VL-7B & 157.06 & 78.53 & 0.75 & 29.3\\
3DThinker\textsubscript{Qwen-2.5VL-7B}* & 157.06 & 78.53 & 0.75 & 63.7\\
VG-LLM-8B* & 218.29 & 109.12 & 1.60 & 62.2\\
Cambrian-\emph{S}(128$\times$64) & 180.88 & 90.44 & 4.22 & 67.5\\
Cambrian-\emph{P}(32$\times$196) & 128.83 & 64.41 & 0.86 & 72.0\\
Cambrian-\emph{P}(64$\times$64) & 80.89 & 40.44 & 1.84 & 73.1\\
Cambrian-\emph{P}(128$\times$64) & 180.88 & 90.44 & 4.22 & 73.7\\
GeoThinker\textsubscript{8frame} & 229.74 & 114.83 & 1.69 & 65.8\\
GeoThinker\textsubscript{32frame} w/o further token compression & 229.74 & 114.83 & 1.69 & --\\
GeoThinker\textsubscript{32frame} & 135.05 & 67.49 & 1.45 & 68.5\\
\rowcolor{navyblue!9}
\name\textsubscript{32frame} w/o further token compression & 173.32 & 86.66 & 1.04 & -- \\
\rowcolor{navyblue!9}
\name\textsubscript{32frame} & 55.33 & 27.66 & 0.48 & 67.3\\


\midrule
\rowcolor{black!9}
\multicolumn{5}{l}{\textbf{Qwen3VL-8B}} \\
Qwen-3VL-8B & 116.95 & 58.47 & 0.62 & 57.7\\
GeoThinker\textsubscript{8frame} & 201.88 & 100.9 & 1.66 & 69.8\\
GeoThinker\textsubscript{32frame} w/o further token compression & 201.88 & 100.9 & 1.66 & --\\
GeoThinker\textsubscript{32frame} & 113.77 & 56.86 & 1.45 & 72.6 \\
\rowcolor{navyblue!9}
\name\textsubscript{32frame} w/o further token compression & 142.59 & 71.29 & 0.91 & -- \\
\rowcolor{navyblue!9}
\name\textsubscript{32frame} & 43.01 & 21.50 & 0.40 & 70.4\\

\bottomrule
\end{tabular}}
\label{tab:appendix_flops}
\end{table}

%% file: tables/21_appendix_recogdrive_latency.tex







\begin{table}[h]
\centering
\caption{
\textbf{Resource efficiency and latency breakdown on autonomous driving evaluation.} We report extra-module and total parameter counts ($M$), computational complexity in FLOPs ($T$), latency breakdowns ($ms$), and amortized Latency (\textit{Avg.}). 
}
\label{tab:appendix_recogdrive_latency}
\setlength{\tabcolsep}{3pt}
\begin{tabular}{l | c c | c c c | c c c c}
& 
\rotatebox{60}{Extra-Module} & 
\rotatebox{60}{Total} & 
\rotatebox{60}{Backbone} & 
\rotatebox{60}{Extra-Module} & 
\rotatebox{60}{Total} & 
\rotatebox{60}{Backbone} & 
\rotatebox{60}{Extra-Modu} & 
\rotatebox{60}{Total} & 
\rotatebox{60}{Avg.} \\

\multirow{-2}{*}{Methods} & 
\multicolumn{2}{c|}{\cellcolor{orange!10}Params (M)} & 
\multicolumn{3}{c|}{\cellcolor{yellow!10}FLOPs (T)} & 
\multicolumn{4}{c}{\cellcolor{blue!10}Latency (ms)} \\

\midrule
ReCogDrive & 0.0 & 7944.4 & 73.45 & 0.00 & 73.45 & 5046.78 & 0 & 5046.78 & 32.15 \\

GeoThinker & 2076.1 & 10020.5 & 73.45 & 49.23 & 122.69 & 5229.24 & 1461.42 & 6690.67 & 42.62 \\

\name & 1187.1 & 9131.6 & 73.45 & 14.77 & 88.22 & 5222.86 & 598.20 & 5821.07 & 37.08 \\
\bottomrule
\end{tabular}
\end{table}

%% file: tables/22_appendix_libero_latency.tex







\begin{table}[h]
\centering
\caption{
\textbf{Computational footprint and inference latency comparison on embodied robotic manipulation.} Parameter size ($M$), FLOPs ($T$), and total latency ($ms$) are decomposed across the backbone and added auxiliary modules. 
}
\label{tab:appendix_libero_latency}
\setlength{\tabcolsep}{4pt}
\begin{tabular}{l | c c | c c c | c}

& 
\rotatebox{60}{Extra-Module} & 
\rotatebox{60}{Total} & 
\rotatebox{60}{Backbone} & 
\rotatebox{60}{Extra-Module} & 
\rotatebox{60}{Total} & 
\\


\multirow{-2}{*}{Methods} & 
\multicolumn{2}{c|}{\cellcolor{orange!10}Params (M)} & 
\multicolumn{3}{c|}{\cellcolor{yellow!10}FLOPs (T)} & 
\cellcolor{blue!10}Latency (ms) \\

\midrule
QwenOFT & 0.0 & 4503.4 & 2.44 & 0.00 & 2.44 & 82.0 \\

GeoThinker & 1705.5 & 6208.9 & 2.44 & 3.40 & 5.84 & 254.8 \\

\name & 277.0 & 4780.4 & 2.44 & 0.07 & 2.51 & 104.2 \\
\bottomrule
\end{tabular}
\end{table}

%% file: tables/11_vstibench.tex
\definecolor{oai-white}{HTML}{FFFFFF}
\definecolor{oai-black}{HTML}{000000}
\definecolor{oai-red}{HTML}{FF4500}
\definecolor{oai-green}{HTML}{51DA4C}
\definecolor{oai-blue}{HTML}{0000FF}
\definecolor{oai-yellow}{HTML}{FFF639}
\definecolor{oai-magenta}{HTML}{FF45FF}
\definecolor{oai-cyan}{HTML}{00FFFF}
\definecolor{oai-orange}{HTML}{FE7600}
\definecolor{oai-violet}{HTML}{8A2BE2}
\definecolor{oai-brown}{HTML}{A0522D}
\definecolor{oai-green-050}{HTML}{F4FFF4}
\definecolor{oai-green-100}{HTML}{E9FFE8}
\definecolor{oai-green-200}{HTML}{D9FFD8}
\definecolor{oai-green-300}{HTML}{C9FFC7}
\definecolor{oai-green-400}{HTML}{A6FFA3}
\definecolor{oai-green-500}{HTML}{7CF178}
\definecolor{oai-green-600}{HTML}{51DA4C}
\definecolor{oai-green-700}{HTML}{3FA93B}
\definecolor{oai-green-800}{HTML}{2D712A}
\definecolor{oai-green-900}{HTML}{193718}
\definecolor{oai-gray-000}{HTML}{FFFFFF}
\definecolor{oai-gray-100}{HTML}{FAFAFA}
\definecolor{oai-gray-200}{HTML}{F5F5F5}
\definecolor{oai-gray-300}{HTML}{E5E5E5}
\definecolor{oai-gray-400}{HTML}{FFB7A4}
\definecolor{oai-gray-500}{HTML}{CDCDCD}
\definecolor{oai-gray-600}{HTML}{A8A8A8}
\definecolor{oai-gray-700}{HTML}{747474}
\definecolor{oai-gray-800}{HTML}{393939}
\definecolor{oai-gray-900}{HTML}{000000}

\begin{table}[htb!]  
    \captionsetup{type=table}
    \centering
    \centering
    \fontsize{4.6pt}{4.4pt}\selectfont 
    \setlength\tabcolsep{3pt}  
    \renewcommand{\arraystretch}{1.4}%
    \caption{
    \textbf{Performance comparison on VSTI-Bench.} GeoThinker-F$_{\text{Qwen3VL-8B}}$ achieves a competitive average score of 64.0\%, outperforming several proprietary and open-source VLM baselines while eliminating the runtime external geometry encoder. Bold and underlined values denote the best and second-best results, respectively.
    }
    \scalebox{1.8}{
    \begin{tabular}{r|c|ccccc} 
    & & 
    \rotatebox{75}{Cam-Obj Abs. Dist.} & 
    \rotatebox{75}{Cam. Displace.} &     
    \rotatebox{75}{Cam. Mov. Dir.} &     
    \rotatebox{75}{Obj-Obj Rel. Pos.} &  
    \rotatebox{75}{Cam-Obj Rel. Dist.} \\ 
    Methods & Avg. &
    \multicolumn{2}{c}{\cellcolor{orange!10}Numerical Answer} & 
    \multicolumn{3}{c}{\cellcolor{yellow!10}Multiple-Choice Answer} \\ 
    \hline
    \rowcolor{black!7}
    \multicolumn{7}{l}{\textcolor{black}{\textit{Baseline}}} \\ 
    Chance Level (Random) & - & - & - & 36.1 & 50.0 & 36.1 \\
    Chance Level (Frequency) & 27.4 & 5.4 & 6.2 & 40.7 & 52.2 & 32.4 \\ 
    \hline
    \rowcolor{black!7}
    \multicolumn{7}{l}{\textcolor{black}{\textit{Human Performance}}} \\
    Human Level & 77.0 & 51.4 & 46.8 & 95.1 & 97.5 & 94.3 \\
    \hline
    \rowcolor{black!7}
    \multicolumn{7}{l}{\textcolor{black}{\textit{Proprietary Models (API)}}} \\
    GPT-4o & 38.2 & 29.5 & 23.4 & 37.3 & 58.1 & 42.5 \\ 
    Gemini-1.5 Flash & 32.1 & 28.5 & 20.9 & 24.4 & 52.6 & 33.9 \\
    \hline
    \rowcolor{black!7}
    \multicolumn{7}{l}{\textcolor{black}{\textit{Open-sourced VLMs}}} \\
    LLaVA-OneVision-0.5B& 36.9 & 16.5 & 32.4 & 46.1 & 50.5 & 39.0 \\
    InternVL2-2B & 38.1 & 17.7 & 27.8 & 43.0 & 54.9 & 47.2 \\
    \hline 
    LLaVA-NeXT-Video-7B & 40.0 & 28.2 & 1.8 & 49.8 & 64.7 & 55.6 \\
    LLaVA-OneVision-7B& 41.7 & 29.9 & 19.3 & 47.5 & 62.1 & 49.8 \\    
    LongVA-7B& 32.3 & 13.5 & 5.1 & 43.7 & 57.9 & 41.2 \\
    InternVL2-8B  & 43.5 & 32.9 & 13.5 & 48.0 & 68.0 & 55.0 \\    
    LongVILA-8B  & 30.5 & 20.0 & 11.6 & 35.4 & 52.3 & 33.4 \\
    VILA-1.5-8B & 37.3 & 30.1 & 27.3 & 42.2 & 50.4 & 36.7 \\
    VILA-1.5-40B & 38.2 & 28.2 & 15.7 & 28.8 & 65.4 & 53.0 \\
    LLaVA-NeXT-Video-72B & 44.0 & 32.3 & 10.5 & 48.1 & 78.3 & 50.9 \\    
    VLM-3R-7B & 58.8 & 39.4 & 39.6 & 60.6 & 86.5 & 68.6 \\
    Cambrian-P & \sota{68.9} & \sota{42.5} & \sota{46.6} & \sota{87.7} & \sota{94.3} & 73.2 \\

    GeoThinker\textsubscript{Qwen3VL-8B} & \underline{67.4} & 38.4 & \underline{45.8} & \underline{84.2} & \underline{93.6} & \sota{75.2} \\
    
    \hline
    \rowcolor{navyblue!10}
    \multicolumn{7}{l}{\textcolor{black}{\textit{Ours}}} \\


    \name\textsubscript{Qwen3VL-8B} & 64.0 & \underline{40.4} & 37.4 & 74.6 & 93.1 & \underline{74.1} \\
    
    \hline
    \end{tabular}
    } 
\label{tab:vsti} 
    \label{tab:appendix_vstibench}
\end{table}

%% file: tables/19_EASI_8_leaderboard.tex
\begin{table*}[ht]
  \caption{\textbf{EASI-8 leaderboard.} 
  VSI denotes VSI-Bench~\citep{yang2025thinking(vsi)}. MMSI denotes MMSI-Bench~\citep{yang2026mmsi}. MindCube denotes MindCube-Tiny~\citep{yin2025spatial(mindcube)}.Viewspatial~\citep{li2025viewspatial},SITE~\citep{wang2025site}, BLINK~\citep{fu2024blink}. 3DSRBench~\citep{ma20253dsrbench}.EmbSpatial denotes EmbSpatial-Bench\citep{du2024embspatial}.
  }
  \label{tab:easi_8_leaderboard}
  \begin{center}
    \begin{small}
      \begin{sc}
      \setlength{\tabcolsep}{3pt}
        \resizebox{\textwidth}{!}{%
        \begin{tabular}{l|cc|cccccccc}
          \toprule

          \multirow{2}{*}{\textbf{Models}}
          & \multirow{2}{*}{\textbf{RANK}}
          & \multirow{2}{*}{\textbf{AVG.}}
          & \multirow{2}{*}{\textbf{VSI}}
          & \multirow{2}{*}{\textbf{MMSI}}
          & \multirow{2}{*}{\textbf{MindCube}}
          & \multirow{2}{*}{\textbf{ViewSpatial}}
          & \multirow{2}{*}{\textbf{SITE}}
          & \multirow{2}{*}{\textbf{BLINK}}
          & \multirow{2}{*}{\textbf{3DSRBench}}
          & \multirow{2}{*}{\textbf{EmbSpatial}}\\

          &&&&&& \\
          
        \midrule
        SenseNova-SI-1.3-InternVL3-8B & 1 & 65.2 & 68.6 & 42.5 & 89.9 & 61.3 & 47.5 & 68.0 & 62.4 & 81.0 \\
        
        \rowcolor{navyblue!13}
        \name\textsubscript{Qwen3VL-8B} & 2 & 64.8 & 70.4 & 30.8 & 93.2 & 82.6 & 53.6 & 55.8 & 55.3 & 76.5 \\

        SenseNova-SI-1.2-InternVL3-8B & 3 & 64.5 & 69.6 & 42.6 & 89.0 & 58.8 & 49.0 & 69.4 & 60.1 & 77.7 \\
        SenseNova-SI-1.5-InternVL3-8B & 4 & 64.4 & 67.3 & 38.3 & 92.1 & 59.0 & 47.5 & 69.5 & 61.1 & 80.4 \\
        SenseNova-SI-1.4-InternVL3-8B & 5 & 63.9 & 66.6 & 40.1 & 88.8 & 55.7 & 47.9 & 68.1 & 62.2 & 81.7 \\
        Gemini 3 Pro& 6 & 63.8 & 52.5 & 45.2 & 70.9 & 50.4 & 62.3 & 76.0 & 68.9 & 84.3 \\
        NEO-ov-9B-SFT & 7 & 63.4 & 64.8 & 41.3 & 90.0 & 55.2 & 54.0 & 62.8 & 61.2 & 77.8 \\
        Qwen3.5-VL-35B-A3B& 8 & 61.6 & 61.1 & 41.9 & 63.5 & 50.8 & 61.4 & 70.5 & 59.8 & 83.6 \\
        SenseNova-SI-1.1-InternVL3-8B & 9 & 61.5 & 68.8 & 43.3 & 85.7 & 54.7 & 47.7 & 63.9 & 55.5 & 72.0 \\
        SenseNova-U1-A3B-MoT& 10 & 59.6 & 56.9 & 39.0 & 70.9 & 58.5 & 47.6 & 59.8 & 63.0 & 81.4 \\
        GPT-5.2& 11 & 59.5 & 54.5 & 41.3 & 60.4 & 47.3 & 61.9 & 71.2 & 60.2 & 79.3 \\
        \rowcolor{black!13}
        GeoThinker\textsubscript{Qwen3VL-8B} & 12 & 59.1 & 72.6 & 30.9 & 83.0 & 45.9 & 55.9 & 53.9 & 51.9 & 78.8 \\
        GPT-5& 13 & 58.8 & 55.0 & 41.8 & 56.3 & 45.6 & 61.9 & 68.0 & 60.3 & 81.6 \\
        SenseNova-U1-8B-MoT& 14 & 58.6 & 57.5 & 38.0 & 61.8 & 55.8 & 49.3 & 63.3 & 63.4 & 79.3 \\
        SenseNova-SI-1.1-Qwen3-VL-8B& 15 & 58.1 & 64.8 & 38.1 & 73.8 & 51.2 & 49.6 & 61.9 & 53.2 & 72.5 \\
        Gemini 2.5 Pro& 16 & 58.0 & 53.6 & 38.0 & 57.6 & 46.1 & 57.1 & 73.5 & 59.3 & 78.8 \\
        Qwen3.5-9B& 17 & 56.1 & 55.7 & 38.7 & 57.6 & 48.2 & 42.3 & 67.6 & 56.8 & 81.5 \\
        NEO-ov-2B-SFT & 18 & 54.4 & 58.4 & 33.6 & 77.2 & 52.8 & 43.0 & 53.8 & 52.5 & 63.8 \\
        Seed 1.6& 19 & 54.2 & 49.9 & 38.3 & 48.8 & 43.9 & 54.6 & 65.9 & 56.9 & 75.4 \\
        Grok 4& 20 & 53.3 & 47.9 & 37.8 & 63.6 & 43.2 & 47.0 & 56.4 & 54.9 & 75.5 \\
        Qwen3-VL-8B-Thinking& 21 & 52.4 & 56.6 & 28.6 & 43.2 & 47.3 & 45.2 & 63.6 & 54.5 & 80.0 \\
        Qwen3-VL-30B-A3B-Thinking& 22 & 52.0 & 51.6 & 29.4 & 40.9 & 47.4 & 47.2 & 63.9 & 55.6 & 80.4 \\
        SenseNova-SI-1.1-Qwen2.5-VL-7B& 23 & 51.0 & 58.1 & 32.8 & 54.7 & 45.5 & 43.9 & 55.3 & 46.3 & 71.4 \\
        Qwen3-VL-8B-Instruct& 24 & 50.8 & 59.8 & 31.1 & 29.4 & 42.2 & 45.8 & 66.7 & 53.9 & 77.7 \\
        VST-7B-SFT & 25 & 50.8 & 55.5 & 32.5 & 39.7 & 50.5 & 39.7 & 61.9 & 53.1 & 73.7 \\
        SenseNova-SI-1.1-InternVL3-2B& 26 & 49.4 & 63.7 & 34.2 & 41.8 & 52.7 & 36.8 & 52.4 & 50.5 & 62.8 \\
        InternVL3\_5-8B& 27 & 49.0 & 56.1 & 29.0 & 40.2 & 40.0 & 43.8 & 58.2 & 49.2 & 75.7 \\
        SenseNova-SI-1.1-BAGEL-7B-MoT& 28 & 48.6 & 41.5 & 34.5 & 46.8 & 46.9 & 42.0 & 65.4 & 42.4 & 69.0 \\
        gemma-4-26B-A4B& 29 & 47.8 & 32.9 & 29.2 & 48.8 & 41.7 & 39.7 & 63.8 & 53.6 & 72.3 \\
        VST-3B-SFT & 30 & 47.7 & 51.4 & 28.8 & 36.0 & 52.9 & 35.9 & 58.8 & 48.7 & 69.0 \\
        vlm-3r-llava-qwen2-lora & 31 & 46.6 & 60.7 & 27.9 & 40.0 & 40.5 & 31.3 & 52.3 & 51.5 & 68.2 \\
        InternVL3-8B& 32 & 45.7 & 42.1 & 28.0 & 41.5 & 38.7 & 41.1 & 53.5 & 44.3 & 76.3 \\
        SenseNova-SI-1.1-Qwen2.5-VL-3B& 33 & 45.7 & 54.9 & 30.8 & 52.6 & 43.5 & 37.8 & 45.6 & 45.0 & 55.3 \\
        BAGEL-7B-MoT& 34 & 45.3 & 31.4 & 31.0 & 34.7 & 41.3 & 37.0 & 63.6 & 50.2 & 73.1 \\
        Qwen3-VL-2B-Instruct& 35 & 45.1 & 54.4 & 28.9 & 34.5 & 37.0 & 35.7 & 53.2 & 47.5 & 70.1 \\
        Cambrian-S-7B & 36 & 45.1 & 62.9 & 27.1 & 37.9 & 41.3 & 36.1 & 37.9 & 45.0 & 72.8 \\
        Qwen2.5-VL-7B-Instruct& 37 & 43.8 & 37.8 & 26.8 & 36.0 & 36.9 & 37.6 & 55.9 & 47.5 & 71.8 \\
        ViLaSR & 38 & 43.7 & 44.6 & 30.2 & 35.1 & 35.7 & 38.7 & 51.4 & 46.6 & 67.3 \\
        InternVL3\_5-2B-Instruct& 39 & 43.2 & 51.3 & 25.4 & 42.5 & 38.1 & 35.3 & 50.2 & 41.5 & 61.5 \\
        Cambrian-S-3B & 40 & 42.0 & 56.1 & 27.0 & 38.4 & 41.0 & 31.0 & 37.7 & 41.4 & 63.5 \\
        SpaceR-SFT-7B & 41 & 41.8 & 41.6 & 27.4 & 38.0 & 35.9 & 34.3 & 49.6 & 40.5 & 66.9 \\
        SpatialLadder-3B & 42 & 40.9 & 44.9 & 27.4 & 43.5 & 39.9 & 28.0 & 43.0 & 42.8 & 58.2 \\
        InternVL3-2B& 43 & 39.8 & 33.0 & 26.5 & 37.5 & 32.6 & 30.0 & 50.8 & 47.7 & 60.1 \\
        Qwen2.5-VL-3B-Instruct& 44 & 39.1 & 26.7 & 28.6 & 37.6 & 32.0 & 33.1 & 48.7 & 43.5 & 62.3 \\
        Spatial-MLLM-subset-sft & 45 & 35.6 & 46.3 & 26.1 & 33.5 & 34.7 & 18.0 & 40.5 & 36.2 & 50.0 \\
        MindCube-Qwen2.5VL-RawQA-SFT & 46 & 22.0 & 17.2 & 1.7 & 51.7 & 24.1 & 6.3 & 35.1 & 2.8 & 37.0 \\
          \bottomrule
        \end{tabular}}
      \end{sc}
    \end{small}
  \end{center}
  \vskip -0.1in
\end{table*}

%% file: tables/10_EASI_ALL_leaderboard.tex
\begin{table*}[ht]
  \caption{\textbf{EASI-ALL leaderboard (In-Domain).} 
  VSI denotes VSI-Bench\citep{yang2025thinking(vsi)}. MMSI denotes MMSI-Bench\citep{yang2026mmsi}. MindCube denotes MindCube-Tiny\citep{yin2025spatial(mindcube)}.Viewspatial\citep{li2025viewspatial},SITE\citep{wang2025site}, BLINK\citep{fu2024blink}. 3DSRBench~\citep{ma20253dsrbench}. EmbSpatial denotes EmbSpatial-Bench\citep{du2024embspatial}. SPAR denotes SPAR-Bench\citep{zhang2025flatland(spar)}. MMSI-Video denotes MMSI-Video-Bench\citep{lin2025mmsi(mmsi-video-bench)}. 
  OmniSpatial\citep{jia2026omnispatial}.
  VSI-Debiased\citep{brown2025benchmark}.
  }
  \label{tab:easi_all_leaderboard}
  \begin{center}
    \begin{small}
      \begin{sc}
      \setlength{\tabcolsep}{3pt}
        \resizebox{\textwidth}{!}{%
        \begin{tabular}{l|cc|cccccccccccc}
          \toprule

          \multirow{2}{*}{\textbf{Models}}
          & \multirow{2}{*}{\textbf{RANK}}
          & \multirow{2}{*}{\textbf{AVG.}}
          & \multirow{2}{*}{\textbf{VSI}}
          & \multirow{2}{*}{\textbf{MMSI}}
          & \multirow{2}{*}{\textbf{MindCube}}
          & \multirow{2}{*}{\textbf{ViewSpatial}}
          & \multirow{2}{*}{\textbf{SITE}}
          & \multirow{2}{*}{\textbf{BLINK}}
          & \multirow{2}{*}{\textbf{3DSRBench}}
          & \multirow{2}{*}{\textbf{EmbSpatial}}
          & \multirow{2}{*}{\textbf{MMSI-Video}}
          & \multirow{2}{*}{\textbf{OmniSpatial}}
          & \multirow{2}{*}{\textbf{SPAR}}
          & \multirow{2}{*}{\textbf{VSI-Debiased}}\\

          & & & & & & & & & & & & & & \\
          
        \midrule
        SenseNova-SI-1.5-InternVL3-8B & 1 & 64.4 & 67.3 & 38.3 & 92.1 & 59.0 & 47.5 & 69.5 & 61.1 & 80.4 & -/- & -/- & -/- & -/- \\
        SenseNova-SI-1.4-InternVL3-8B & 2 & 63.9 & 66.6 & 40.1 & 88.8 & 55.7 & 47.9 & 68.1 & 62.2 & 81.7 & -/- & -/- & -/- & -/- \\
        NEO-ov-9B-SFT & 3 & 63.4 & 64.8 & 41.3 & 90.0 & 55.2 & 54.0 & 62.8 & 61.2 & 77.8 & -/- & -/- & -/- & -/- \\
        Qwen3.5-VL-35B-A3B & 4 & 61.6 & 61.1 & 41.9 & 63.5 & 50.8 & 61.4 & 70.5 & 59.8 & 83.6 & -/- & -/- & -/- & -/- \\
        Gemini 3 Pro & 5 & 60.2 & 52.5 & 45.2 & 70.9 & 50.4 & 62.3 & 76.0 & 68.9 & 84.3 & 40.4 & 69.1 & 48.7 & 53.8 \\
        
        \rowcolor{navyblue!13}
        \name\textsubscript{Qwen3VL-8B} & 6 & 60.1 & 70.4 & 30.8 & 93.2 & 82.6 & 53.6 & 55.8 & 55.3 & 76.5 & 28.4 & 42.5 & 66.9 & 65.8\\
        
        SenseNova-U1-A3B-MoT & 7 & 59.6 & 56.9 & 39.0 & 70.9 & 58.5 & 47.6 & 59.8 & 63.0 & 81.4 & -/- & -/- & -/- & -/- \\
        SenseNova-U1-8B-MoT & 8 & 58.6 & 57.5 & 38.0 & 61.8 & 55.8 & 49.3 & 63.3 & 63.4 & 79.3 & -/- & -/- & -/- & -/- \\
        SenseNova-SI-1.3-InternVL3-8B & 9 & 57.8 & 68.6 & 42.5 & 89.9 & 61.3 & 47.5 & 68.0 & 62.4 & 81.0 & 25.7 & 35.3 & 48.4 & 63.0 \\
        SenseNova-SI-1.2-InternVL3-8B & 10 & 57.7 & 69.6 & 42.6 & 89.0 & 58.8 & 49.0 & 69.4 & 60.1 & 77.7 & 26.2 & 34.8 & 49.5 & 65.2 \\
        GPT-5.2 & 11 & 56.3 & 54.5 & 41.3 & 60.4 & 47.3 & 61.9 & 71.2 & 60.2 & 79.3 & 32.3 & 57.4 & 55.1 & 54.5 \\
        Qwen3.5-9B & 12 & 56.1 & 55.7 & 38.7 & 57.6 & 48.2 & 42.3 & 67.6 & 56.8 & 81.5 & -/- & -/- & -/- & -/- \\
        GPT-5 & 13 & 55.4 & 55.0 & 41.8 & 56.3 & 45.6 & 61.9 & 68.0 & 60.3 & 81.6 & 33.4 & 59.2 & 49.7 & 52.5 \\

        \rowcolor{black!13}
        GeoThinker\textsubscript{Qwen3VL-8B} & 14 & 55.0 & 72.6 & 30.9 & 83.0 & 45.9 & 55.9 & 53.9 & 51.9 & 78.8 & 23.7 & 40.1 & 68.2 & -/- \\
        
        SenseNova-SI-1.1-InternVL3-8B & 15 & 54.8 & 68.8 & 43.3 & 85.7 & 54.7 & 47.7 & 63.9 & 55.5 & 72.0 & 23.8 & 33.0 & 45.8 & 62.9 \\
        Gemini 2.5 Pro & 16 & 54.7 & 53.6 & 38.0 & 57.6 & 46.1 & 57.1 & 73.5 & 59.3 & 78.8 & 32.9 & 57.6 & 49.8 & 51.8 \\
        NEO-ov-2B-SFT & 17 & 54.4 & 58.4 & 33.6 & 77.2 & 52.8 & 43.0 & 53.8 & 52.5 & 63.8 & -/- & -/- & -/- & -/- \\
        SenseNova-SI-1.1-Qwen3-VL-8B & 18 & 52.8 & 64.8 & 38.1 & 73.8 & 51.2 & 49.6 & 61.9 & 53.2 & 72.5 & 25.5 & 43.0 & 40.8 & 59.4 \\
        Qwen3-VL-8B-Thinking & 19 & 52.4 & 56.6 & 28.6 & 43.2 & 47.3 & 45.2 & 63.6 & 54.5 & 80.0 & -/- & -/- & -/- & -/- \\
        Qwen3-VL-30B-A3B-Thinking & 20 & 52.0 & 51.6 & 29.4 & 40.9 & 47.4 & 47.2 & 63.9 & 55.6 & 80.4 & -/- & -/- & -/- & -/- \\
        Seed 1.6 & 21 & 51.0 & 49.9 & 38.3 & 48.8 & 43.9 & 54.6 & 65.9 & 56.9 & 75.4 & 29.8 & 50.6 & 50.1 & 48.3 \\
        Grok 4 & 22 & 49.3 & 47.9 & 37.8 & 63.6 & 43.2 & 47.0 & 56.4 & 54.9 & 75.5 & 27.2 & 47.7 & 44.8 & 45.7 \\
        Qwen3-VL-8B-Instruct & 23 & 47.9 & 59.8 & 31.1 & 29.4 & 42.2 & 45.8 & 66.7 & 53.9 & 77.7 & 28.4 & 47.0 & 39.6 & 53.1 \\
        gemma-4-26B-A4B & 24 & 47.8 & 32.9 & 29.2 & 48.8 & 41.7 & 39.7 & 63.8 & 53.6 & 72.3 & -/- & -/- & -/- & -/- \\
        VST-7B-SFT & 25 & 47.2 & 55.5 & 32.5 & 39.7 & 50.5 & 39.7 & 61.9 & 53.1 & 73.7 & 24.9 & 39.5 & 46.6 & 48.8 \\
        SenseNova-SI-1.1-Qwen2.5-VL-7B & 26 & 47.1 & 58.1 & 32.8 & 54.7 & 45.5 & 43.9 & 55.3 & 46.3 & 71.4 & 26.1 & 39.3 & 38.2 & 53.4 \\
        InternVL3\_5-8B & 27 & 46.3 & 56.1 & 29.0 & 40.2 & 40.0 & 43.8 & 58.2 & 49.2 & 75.7 & 28.0 & 47.4 & 38.2 & 50.5 \\
        vlm-3r-llava-qwen2-lora & 28 & 45.1 & 60.7 & 27.9 & 40.0 & 40.5 & 31.3 & 52.3 & 51.5 & 68.2 & 27.8 & 43.3 & 42.4 & 55.5 \\
        SenseNova-SI-1.1-BAGEL-7B-MoT & 29 & 44.7 & 41.5 & 34.5 & 46.8 & 46.9 & 42.0 & 65.4 & 42.4 & 69.0 & 23.8 & 44.0 & 44.7 & 35.0 \\
        SenseNova-SI-1.1-InternVL3-2B & 30 & 44.6 & 63.7 & 34.2 & 41.8 & 52.7 & 36.8 & 52.4 & 50.5 & 62.8 & 20.4 & 26.4 & 38.0 & 56.0 \\
        VST-3B-SFT & 31 & 43.9 & 51.4 & 28.8 & 36.0 & 52.9 & 35.9 & 58.8 & 48.7 & 69.0 & 24.3 & 36.5 & 37.7 & 46.5 \\
        Cambrian-S-7B & 32 & 43.5 & 62.9 & 27.1 & 37.9 & 41.3 & 36.1 & 37.9 & 45.0 & 72.8 & 25.2 & 41.9 & 37.9 & 55.4 \\
        InternVL3-8B & 33 & 43.0 & 42.1 & 28.0 & 41.5 & 38.7 & 41.1 & 53.5 & 44.3 & 76.3 & 30.2 & 45.3 & 35.9 & 38.5 \\
        SenseNova-SI-1.1-Qwen2.5-VL-3B & 34 & 41.8 & 54.9 & 30.8 & 52.6 & 43.5 & 37.8 & 45.6 & 45.0 & 55.3 & 25.1 & 32.5 & 30.8 & 47.8 \\
        BAGEL-7B-MoT & 35 & 41.7 & 31.4 & 31.0 & 34.7 & 41.3 & 37.0 & 63.6 & 50.2 & 73.1 & 27.8 & 41.7 & 39.1 & 29.0 \\
        Qwen3-VL-2B-Instruct & 36 & 41.6 & 54.4 & 28.9 & 34.5 & 37.0 & 35.7 & 53.2 & 47.5 & 70.1 & 26.6 & 34.6 & 33.9 & 43.3 \\
        InternVL3\_5-2B-Instruct & 37 & 40.9 & 51.3 & 25.4 & 42.5 & 38.1 & 35.3 & 50.2 & 41.5 & 61.5 & 25.4 & 41.1 & 32.5 & 45.7 \\
        Cambrian-S-3B & 38 & 40.3 & 56.1 & 27.0 & 38.4 & 41.0 & 31.0 & 37.7 & 41.4 & 63.5 & 23.9 & 41.9 & 33.0 & 48.8 \\
        Qwen2.5-VL-7B-Instruct & 39 & 39.8 & 37.8 & 26.8 & 36.0 & 36.9 & 37.6 & 55.9 & 47.5 & 71.8 & 27.1 & 37.4 & 33.8 & 28.9 \\
        ViLaSR & 40 & 39.4 & 44.6 & 30.2 & 35.1 & 35.7 & 38.7 & 51.4 & 46.6 & 67.3 & 28.3 & 19.2 & 37.4 & 38.3 \\
        SpaceR-SFT-7B & 41 & 38.8 & 41.6 & 27.4 & 38.0 & 35.9 & 34.3 & 49.6 & 40.5 & 66.9 & 24.7 & 41.0 & 34.2 & 31.6 \\
        SpatialLadder-3B & 42 & 38.7 & 44.9 & 27.4 & 43.5 & 39.9 & 28.0 & 43.0 & 42.8 & 58.2 & 27.4 & 41.9 & 32.9 & 35.2 \\
        InternVL3-2B & 43 & 37.1 & 33.0 & 26.5 & 37.5 & 32.6 & 30.0 & 50.8 & 47.7 & 60.1 & 29.1 & 42.0 & 27.2 & 28.6 \\
        Qwen2.5-VL-3B-Instruct & 44 & 36.1 & 26.7 & 28.6 & 37.6 & 32.0 & 33.1 & 48.7 & 43.5 & 62.3 & 27.7 & 41.1 & 28.3 & 23.5 \\
        Spatial-MLLM-subset-sft & 45 & 34.9 & 46.3 & 26.1 & 33.5 & 34.7 & 18.0 & 40.5 & 36.2 & 50.0 & 22.6 & 38.0 & 35.3 & 38.0 \\
        MindCube-Qwen2.5VL-RawQA-SFT & 46 & 20.4 & 17.2 & 1.7 & 51.7 & 24.1 & 6.3 & 35.1 & 2.8 & 37.0 & 5.2 & 24.5 & 20.8 & 18.6 \\
          \bottomrule
        \end{tabular}}
      \end{sc}
    \end{small}
  \end{center}
  \vskip -0.1in
\end{table*}

%% file: tables/14_appendix_revsi.tex
\begin{table*}[ht]
  \caption{
  \textbf{Performance comparisons on ReVSI-Bench.} ReVSI~\citep{zhang2026revsi} introduces a dynamic evaluation setup where ground-truth answers adapt according to the evaluated frame count. 
  \name maintains competitive accuracy on ReVSI-Bench without executing a runtime external geometry encoder, while trailing its external-encoder GeoThinker counterpart.
  }
  \label{tab:appendix_revsi}
  \begin{center}
    \begin{small}
      \begin{sc}
      \setlength{\tabcolsep}{3pt}
        \resizebox{\textwidth}{!}{%
        \begin{tabular}{lc|c|cccc|ccc}
          \toprule
          \multirow{2}{*}{Methods}& \multirow{2}{*}{Frames} & \multirow{2}{*}{Avg.}
          & \multicolumn{4}{c|}{\cellcolor{orange!10}Numerical Answer}
          & \multicolumn{3}{c}{\cellcolor{yellow!10}Multiple-Choice Answer} \\
          \cmidrule(lr){4-7}\cmidrule(lr){8-10}
          & &&  Obj.Count & Abs. Dist. & Obj. Size & Room Size & Rel. Dist. & Rel. Dir. & Route Plan  \\
          \midrule
          \rowcolor{black!11}
          \multicolumn{10}{l}{\textbf{Proprietary Models}} \\
          GPT-5.2 & 64 & 50.9 & 56.2 & 41.5 & 73.9 & \textbf{63.0} & 48.4 & 34.9 & 38.2 \\
          Gemini 3 Flash & 1 FPS & 57.6 & \textbf{65.7} & 53.1 & 77.6 & 52.8 & 64.6 & 47.9 & 41.8 \\
          Gemini 3 Pro & 1 FPS & \textbf{60.9} & 60.1 & 54.7 & \textbf{79.3} & 51.9 & \textbf{68.1} & \textbf{56.0} & \textbf{56.4} \\

          \midrule
          \rowcolor{black!8}
          \multicolumn{10}{l}
          {\textbf{Open-source Models}} \\
          Qwen3-VL-8B-Instruct~\citep{Qwen3-VL} & 64 & 49.1 & 40.4 & 52.3 & \textbf{69.0} & 45.1 & 57.1 & 39.5 & 40.5 \\
          Qwen2.5-VL-7B-Instruct~\citep{Qwen2.5-VL} & 4 FPS & 32.6 & 36.9 & 15.0 & 49.7 & 29.0 & 31.5 & 29.5 & 36.7 \\
          InternVL3.5-8B~\citep{wang2025internvl3.5} & 64 & 47.9 & 43.3 & 54.6 & 64.2 & 47.6 & 45.0 & 36.3 & 44.4 \\
          LLaVA-Video-7B-Qwen2~\citep{zhang2024video} & 64 & 30.3 & 31.3 & 1.4 & 52.5 & 16.7 & 38.3 & 33.3 & 38.4 \\
          Cambrian-\emph{S}\xspace-7B~\citep{yang2025cambrian} & 128 & 49.1 & \textbf{48.4} & 60.5 & 65.5 & 46.7 & 37.1 & 48.5 & 37.0 \\
          VST-7B-SFT~\citep{vst2025} & 4 FPS & 46.4 & 35.4 & 52.6 & 67.9 & 47.2 & 49.2 & 36.9 & 35.4 \\
          Cambrian-\emph{P}\xspace\ (w/o Pose) & 128 & 50.1 & 42.3 & 64.6 & 64.5 & 47.4 & 38.1 & 48.3 & 45.3 \\
          Cambrian-\emph{P}\xspace~\citep{yang2026cambrian(cambrianP)} & 128 & 52.0 & 41.4 & 68.3 & 66.7 & 45.9 & 40.2 & \textbf{48.7} & \textbf{53.1} \\

          GeoThinker\textsubscript{Qwen3VL-8B}~\citep{li2026thinking(geothinker)} & 128 & \sota{55.9} & 37.9 & \sota{74.7} & 67.6 & \sota{57.1} & \sota{61.8} & 47.8 & 44.7 \\

        \midrule
          \rowcolor{navyblue!11}
          \multicolumn{10}{l}{\textbf{Ours}} \\
          
          
          \name\textsubscript{Qwen3VL-8B} & 128 & 50.6 & 36.4 & 65.2 & 66.1 & 56.2 & 45.9 & 46.1 & 38.5 \\

          \bottomrule
        \end{tabular}%
        }
      \end{sc}
    \end{small}
  \end{center}
  \vskip -0.2in
\end{table*}

%% file: tables/15_appendix_revis_frame.tex
\begin{table}[t]
\centering
\small
\setlength{\tabcolsep}{1.5mm}
\caption{
\textbf{Zero-shot frame scaling on ReVSI.} Evaluates temporal extrapolation from 16 to 128 frames, where target ground-truth annotations dynamically adjust to the changing visual context.
}
\resizebox{0.7\linewidth}{!}{
\begin{tabular}{llcccc}
\toprule
\multirow{2}{*}{\textbf{Model}} & \multirow{2}{*}{\textbf{Benchmark}} & \multicolumn{4}{c}{\textbf{\# Frames}} \\
\cmidrule(lr){3-6}
 &  & \textbf{16} & \textbf{32} & \textbf{64} & \textbf{128} \\

\midrule
\multirow{3}{*}{GeoThinker\textsubscript{Qwen3vl-8B-32frame}}
& VSI & 69.2 & 72.6 & 73.4 & \sota{73.4} \\
& VSI-Debiased & 64.3 & 66.3 & 67.7 & \sota{68.1} \\
& ReVSI & 52.6 & 53.9 & 54.8 & \sota{55.9}\\

\midrule
\multirow{3}{*}{\name\textsubscript{Qwen3vl-8B-32frame}}
& VSI & 67.0 & 70.4 & 72.1 & \sota{72.4} \\
& VSI-Debiased & 61.7 & 65.8 & \sota{67.1} & 66.7 \\
& ReVSI & 48.6 & 49.8 & 50.4 & \sota{50.6} \\
 
\bottomrule
\end{tabular}}
\label{tab:appendix_revis_frame}
\end{table}

%% file: tables/16_appendix_all_no_omnispatial.tex
\begin{table*}[ht]
  \caption{
  \textbf{Ablation study on the influence of the OmniSpatial-Train data mixture.} Controlled comparison between GeoThinker and \name with and without the OmniSpatial training dataset, isolating the core architectural contributions of endogenous feature prediction under strictly aligned training data distributions.
  }
  \vskip -0.1in
  \label{tab:appendix_no_omnispatial}
  \begin{center}
    \begin{small}
      \begin{sc}
      \setlength{\tabcolsep}{3pt}
        \resizebox{\textwidth}{!}{%
        \begin{tabular}{l|c|c|cccccc}
          \toprule


          \multirow{2}{*}{\textbf{Models}}
          & \textbf{Geometry}
          & \multirow{2}{*}{\textbf{AVG.}}
          & \multirow{2}{*}{\textbf{VSI-Bench}}
          & \multirow{2}{*}{\textbf{MMSI-Bench}}
          & \multirow{2}{*}{\textbf{MindCube}}
          & \multirow{2}{*}{\textbf{ViewSpatial}}
          & \multirow{2}{*}{\textbf{SITE}}
          & \multirow{2}{*}{\textbf{CV-Bench}}\\

          & \textbf{Encoder} & & & & & & & \\

          \midrule
          \rowcolor{navyblue!9}
          \multicolumn{9}{l}{\textbf{Qwen2.5VL-7B}} \\
          
          GeoThinker & \cmarkg & 60.43 & 68.5 & 31.7 & 83.6 & 41.9 & 54.8 & 82.1 \\

          \name & \xmark & 60.56 & 67.7 & 31.0 & 88.3 & 42.1 & 53.6 & 80.5 \\

          \name + OmniSpatial-Train & \xmark & \sota{66.84} & 67.3 & 31.2 & 86.8 & 83.2\textcolor{blue}{\textbf{(+41.1)}} & 53.2 & 79.1 \\

            \midrule
            \rowcolor{navyblue!9}
            \multicolumn{9}{l}{\textbf{Qwen3VL-8B}} \\
            GeoThinker & \cmarkg & 62.23 & 72.6 & 30.9  & 83.0  & 45.9 & 55.9 & 85.1  \\

            \name & \xmark & 62.21 & 70.9 & 29.6 & 91.0 & 44.7 & 54.2 & 82.9 \\
            
            \name + OmniSpatial-Train & \xmark & \sota{69.29} & 70.4 & 30.8  & 93.2  & 82.6\textcolor{blue}{\textbf{(+37.9)}} & 53.7 & 85.0  \\

          \bottomrule
        \end{tabular}}
      \end{sc}
    \end{small}
  \end{center}
  \vspace{-5mm}
\end{table*}

%% file: tables/17_appendix_stage3_ablation.tex
\begin{table}[ht]
  \caption{
  \textbf{Ablation study on Stage-2 training strategies for \name.} Evaluated on a Qwen2.5-VL-7B backbone across frame budgets (8 vs. 32 frames), General Video mixture integration, and LFP head fine-tuning status. Orange numbers at ($\cdot$) denote performance gaps relative to GeoThinker. Jointly fine-tuning LFP heads alongside general video mixtures yields the optimal balance across specialized spatial and multi-modal benchmarks.
  }
  \vskip -0.1in
  \label{tab:appendix_stage2_ablation}
  \begin{center}
  \setlength{\tabcolsep}{2pt}
    \begin{small}
      \begin{sc}
        \resizebox{0.8\linewidth}{!}{%
        \begin{tabular}{ccc|ccc|c}
          \toprule


        Training & General & LFP & \multirow{2}{*}{VSI$\uparrow$} & \multirow{2}{*}{MMSI$\uparrow$} & Cross-Bench. & \multirow{2}{*}{Latency$\downarrow$}\\

         Frame & Video & FineTuning & & & Avg.$\uparrow$ & \\

          \midrule
          \rowcolor{black!9}
          \multicolumn{7}{l}{\textbf{GeoThinker}} \\
          
           8 & \xmarkg & -- & 65.8 & 29.1 & 57.67 & 1.69\\

           32 & \cmarkg & -- & 68.5 & 31.7 & 60.43 & 1.45\\

          \midrule
          \rowcolor{navyblue!9}
          \multicolumn{7}{l}{\textbf{\name}} \\

           8 & \xmarkg & \xmarkg & 63.6\textcolor{orange}{(-2.2)} & 28.1\textcolor{orange}{(-1.0)} & 65.42 & 1.04\\

           \midrule

            32 & \xmarkg & \xmarkg & 65.2 & 27.9 & 65.55 & \multirow{4}{*}{0.48}\\
           

            32 & \cmark & \xmarkg & 67.6 & 28.1 & 67.15 & \\

            32 & \xmarkg & \cmark & 66.8 & 28.9 & 66.47 & \\

            32 & \cmark & \cmark & 67.3\textcolor{orange}{(-1.2)} & 31.2\textcolor{orange}{(-0.5)} & 66.84 & \\

          \bottomrule
        \end{tabular}
        }
      \end{sc}
    \end{small}
  \end{center}
  \vspace{-5mm}
\end{table}